\documentclass[acmsmall]{acmart}

\usepackage{algorithm}
\usepackage[noend]{algorithmic}
\usepackage{multicol}
\usepackage{multirow}
\usepackage{xfrac}
\usepackage{caption}
\usepackage{subcaption}
\usepackage{booktabs}
\usepackage{pifont}
\usepackage{enumitem}
\usepackage{textcomp}
\usepackage[xcolor]{changebar}
\usepackage{dsfont}

\AtBeginDocument{%
  \providecommand\BibTeX{{%
    \normalfont B\kern-0.5em{\scshape i\kern-0.25em b}\kern-0.8em\TeX}}}

\newcommand{\halfquad}{\hspace{0.5em}} 

\setcopyright{acmcopyright}
\copyrightyear{2022}
\acmYear{2022}
\acmDOI{10.1145/3524885}

\acmJournal{TIST}
\acmVolume{13}
\acmNumber{4}
\acmArticle{78}
\acmMonth{3}

\newcommand\numberthis{\addtocounter{equation}{1}\tag{\theequation}}

\begin{document}

\title[Semi-Synchronous Federated Learning]{Semi-Synchronous Federated Learning for Energy-Efficient Training and Accelerated Convergence in Cross-Silo Settings}

\author{Dimitris Stripelis}
\email{stripeli@isi.edu}
\orcid{}
\affiliation{%
  \institution{Information Sciences Institute, University of Southern California}
  \streetaddress{4676 Admiralty Way}
  \city{Marina Del Rey}
  \state{CA}
  \country{USA}
  \postcode{90292}
}

\author{Paul M. Thompson}
\email{thompson@ini.usc.edu}
\orcid{}
\affiliation{%
 \institution{Imaging Genetics Center, Stevens Neuroimaging and Informatics Institute, University of Southern California}
  \streetaddress{4676 Admiralty Way}
  \city{Marina Del Rey}
  \state{CA}
  \country{USA}
  \postcode{90292}
}

\author{Jos\'{e} Luis Ambite}
\email{ambite@isi.edu}
\orcid{}
\affiliation{%
  \institution{Information Sciences Institute, University of Southern California}
  \streetaddress{4676 Admiralty Way}
  \city{Marina Del Rey}
  \state{CA}
  \country{USA}
  \postcode{90292}
}

\renewcommand{\shortauthors}{Stripelis et al.}

\begin{abstract}
There are situations where data relevant to machine learning problems are distributed across multiple locations that cannot share the data due to regulatory, competitiveness, or privacy reasons. Machine learning approaches that require data to be copied to a single location are hampered by the challenges of data sharing. Federated Learning (FL) is a promising approach to learn a joint model over all the available data across silos. In many cases, the sites participating in a federation have different data distributions and computational capabilities. In these heterogeneous environments existing approaches exhibit poor performance: synchronous FL protocols are communication efficient, but have slow learning convergence and high energy cost; conversely, asynchronous FL protocols have faster convergence with lower energy cost, but higher communication. In this work, we introduce a novel energy-efficient \textit{Semi-Synchronous Federated Learning} protocol that mixes local models periodically with minimal idle time and fast convergence. We show through extensive experiments over established benchmark datasets in the computer-vision domain as well as in real-world biomedical settings that our approach significantly outperforms previous work in \textit{data and computationally heterogeneous environments}. 
\end{abstract}




\maketitle

\section{Introduction}
Data useful for a machine learning problem is often generated at multiple, distributed locations. In many situations these data cannot be exported from their original location due to regulatory, competitiveness, or privacy reasons. A primary motivating example is health records, which are heavily regulated and protected, restricting the ability to analyze large datasets. Industrial data (e.g., accident or safety data) is also not shared due to competitiveness reasons. Additionally, given recent high-profile data leak incidents more strict data ownership laws have been enacted, such as the European Union's General Data Protection Regulation (GDPR), China's Cyber Security Law and General Principles of Civil Law, and the California Consumer Privacy Act (CCPA). 

These situations bring data distribution, security, and privacy to the forefront of machine learning and impose new challenges on how data should be shared and analyzed. Federated Learning (FL) is a promising solution~\cite{mcmahan2017communication,konevcny2016federated,yang2019federated}, which can help learn deep neural networks from data silos~\cite{jain2003out} by collaboratively training models that aggregate locally-computed updates (e.g., gradients) under a centralized (e.g., central parameter server) or a decentralized (e.g., peer-to-peer) learning topology~\cite{li2020federatedopt,bellavista2021decentralised}, while providing strong privacy and security guarantees (e.g.,~\cite{bonawitz2017practical,zhang2020batchcrypt}). 
Models trained using Federated Learning outperform the models that any individual participant in the federation could achieve by training solely on its local data (cf. Figure~\ref{fig:Silos_vs_Federation}).

Our primary interest is to develop efficient Federated Learning training policies for the data center setting~\cite{li2020federatedopt} that accelerate learning convergence while minimizing computational, communication, and energy costs. In particular, we target cross-silo Federated Learning environments~\cite{kairouz2019advances,yang2019federated,li2020federatedopt,rieke2020future,zhang2020batchcrypt}, 
such as a network of hospitals or a research consortium that wants to jointly learn a model without sharing any data. In these environments, participants often have different computational capabilities (\textit{system heterogeneity}) and different local data distributions (\textit{statistical heterogeneity}). Current Federated Learning approaches use either synchronous~\cite{mcmahan2017communication,smith2017federated,bonawitz2019towards} or asynchronous~\cite{xie2019asynchronous} communication protocols. However, these methods have poor performance in such heterogeneous environments. Synchronous protocols result in computationally fast learners being idle, underutilizing the available resources of the federation. Asynchronous protocols fully utilize the available resources, but incur higher network communication cost and potentially lower generalizability due to the effect of stale models~\cite{cui2014exploiting,dai2018toward}.

To address these challenges, we introduce a new hybrid training scheme, \textit{Semi-Synchronous Federated Learning (SemiSync)}, which allows learners to continuously train on their local dataset up to a specific temporally specified synchronization point where the current local models of all learners are aggregated to compute the community model. This approach produces better utilization of the federation resources, while limiting communication costs. 
We empirically demonstrate the effectiveness of this new scheme in terms of convergence time, communication cost and energy efficiency on a variety of challenging learning environments with diverse computational resources, variable data amounts, and different target class distributions (IID and Non-IID example assignments) across learners. 
Our training scheme leads to faster convergence regardless of the optimizer used to train the local models. 
Our contributions are:
\begin{itemize}[noitemsep, topsep=5pt]
    \item A new \textit{Semi-Synchronous} communication protocol for cross-silo Federated Learning settings with accelerated convergence, and reduced communication, processing and energy cost. %
    \item A caching approach that enables computation of the community model in constant time in asynchronous Federated Learning environments.
    \item A new asynchronous communication policy based on staleness, \textit{FedRec}, which outperforms existing asynchronous learning policies.
    \item A systematic evaluation of Federated Learning training policies on standard benchmarks and on a challenging neuroimaging dataset over diverse federated learning environments.
\end{itemize}

\section{Related Work}\label{sec:RelatedWork}

Federated Learning was introduced by McMahan et al. (\citeyear{mcmahan2017communication}) for user data in mobile phones. Their original algorithm, \textit{Federated Average}, follows a synchronous communication protocol where each learner (phone) trains a neural network for a fixed number of epochs on its local dataset. Once all learners (or a subset) finish their locally assigned training, the system (i.e., federation controller) computes a community model that is a weighted average of each of the learners' local models, with the weight of each learner in the federation being equal to the number of its local training examples. The new community model is then distributed to all learners and the process repeats. This approach, which we call \textit{SyncFedAvg}, has catalyzed much of the recent work~\cite{smith2017federated,bonawitz2019towards,li2020federatedopt}.

\paragraph{SGD Optimization.} The FL settings that we investigate in this work are related to stochastic optimization in distributed and parallel systems~\cite{bertsekas1983distributed,bertsekas1989parallel}, as well as in synchronous distributed stochastic gradient descent (SGD) optimization~\cite{chen2016revisiting}. The problem of delayed (i.e., stale) gradient updates due to asynchronicity is well known~\cite{lian2015asynchronous,recht2011hogwild,agarwal2011distributed}, with~\cite{lian2015asynchronous} providing theoretical support for nonconvex optimization functions under the IID assumption. In our work, we study FL in more general, Non-IID settings.

\paragraph{Periodic SGD} Our SemiSync policy is related to parallel training policies such as Elastic and Periodic SGD Averaging
\cite{zhang2015deep,wang2018cooperative,reisizadeh2020fedpaq}, where a central server aggregates the learners local gradient updates when a specific number of iterations is complete. This periodic aggregation controls the communication frequency between every learner and the server. 
Zhang et al.~\cite{zhang2015deep} introduce an elastic update rule around an average global variable consensus for IID settings. 
Wang and Joshi~\cite{wang2018cooperative} provide a general analysis framework for different cooperative learning environments. 
FedPAQ~\cite{reisizadeh2020fedpaq} studies the convergence guarantees of periodic averaging in cross-device IID settings for strongly convex and non-convex functions with quantized message-passing. 
In contrast, we study \textit{time-based} periodic averaging when the participating learners have highly diverse computational resources and data distributions. 
We define the synchronization time period based on the data amounts and computational power of the learners, as multiples/fractions of the time it takes for the slowest learner in the federation to complete a single epoch, instead of epoch-level~\cite{mcmahan2017communication} or iteration-level~\cite{reisizadeh2020fedpaq} synchronization, or performance tiers~\cite{chai2020tifl}.

\paragraph{Global and Local Model Optimization.} In statistically heterogeneous FL settings clients can drift too far away from the global optimal model. An approach to tackle client drift is to decouple the SGD optimization into local (learner side) and global (server side)~\cite{reddi2020adaptive,hsu2019measuring}.
Hsu et al.~\cite{hsu2019measuring} investigate a server-side momentum-based update rule (\textsc{FedAvgM}) between the previous community model and the newly computed weighted average of the clients' models.
Reddi et al.~\cite{reddi2020adaptive} introduce a more generalized server-side optimization update rule (\textsc{FedOpt}) with support for adaptive SGD optimizers (\textsc{FedAdagrad, FedYogi, FedAdam}).
FedProx~\cite{li2020federatedopt} directly addresses client drift by introducing a regularization term in the clients local objective, which penalizes the divergence of the local solution from the global solution.
FedAsync~\cite{xie2019asynchronous} weights the different local models based on their staleness with respect to the latest community model.
Liu et al~\cite{liu2020accelerating} show that using Momentum SGD as the local model solver leads to accelerated convergence compared to Vanilla SGD.
In our setting, we study the effect of local models mixing strategies under different communication protocols, as well as the effect of local model solvers (i.e., Vanilla SGD, Momentum SGD and FedProx) on the convergence rate of the global model.

\paragraph{Federated Convergence Guarantees.} 
FedProx studied the convergence rate of FedAvg over dissimilar learners' local solutions~\cite{li2020federatedopt}.
Wang et al.~\cite{wang2019adaptive} studied adaptive FL in mobile edge computing environments under resource budget constraints with arbitrary local updates between learners. Li et al. ~\cite{Li2020On} provide convergence guarantees over full and partial device participation for FedAvg. 
Chen et al.~\cite{chen2020joint} formulated a joint federated learning optimization problem targeting optimal resource allocation and client selection in wireless networks.
FedAsync~\cite{xie2019asynchronous} provides convergence guarantees for asynchronous environments and a community model that is a weighted average of local models based on staleness. 
In our work, we empirically study the convergence of the different federated learning protocols on computationally heterogeneous environments on IID and Non-IID data distributions with full client participation in the cross-silo (data center) FL settings with the presence of stragglers~\cite{dean2013tail}, instead of partial client participation~\cite{li2020federatedopt,mcmahan2017communication}, dropouts~\cite{chai2020tifl}.

\paragraph{Federated Learning Energy Efficiency} Most of the recent work on energy consumption in FL settings focuses on the energy cost of decentralized training on wireless networks~\cite{luo2020cost,yang2020energy,tran2019federated}. Luo et al.~\cite{luo2020cost} consider the problem of cost-effective FL design which jointly optimizes learning time and energy consumption on the edge with convergence guarantees. The works of~\cite{tran2019federated,yang2020energy} study the trade-off of computation and communication latency in decentralized FL and its effect on total energy consumption, system learning time and learning accuracy. In our work, however, we analyze the energy efficiency of federated learning training protocols in cross-silo (data center) settings. As it is also shown in~\cite{shehabi2016united} energy consumption in data centers is of notable importance due to its considerable environmental impact, while the work of~\cite{masanet2020recalibrating} stresses the need for additional energy consumption analysis tools that can monitor more accurately energy efficiency and help advance sustainability in data center settings.

\paragraph{Federated Learning in Healthcare.} Federated Learning holds great promise for the future of digital health and cross-institutional healthcare informatics~\cite{rieke2020future}. 
FL has been used for phenotype discovery~\cite{liu2019two}, for patient representation learning~\cite{kim2017federated}, and for identifying similar patients across institutions~\cite{lee2018privacy}. 
FL has been applied to a variety of tasks in biomedical imaging, including whole-brain segmentation of MRI T1 scans~\cite{roy2019braintorrent}, brain tumor segmentation~\cite{sheller2018multi,li2019privacy}, multi-site fMRI classification and identification of disease biomarkers~\cite{li2020multi}, and for identification of brain structural relationships across diseases and clinical cohorts using (federated) dimensionality reduction from shape features~\cite{silva2019federated}. 
Silva et al.~\cite{silva2020fed} present an open-source FL framework for healthcare that can support different machine learning models and optimization methods.
COINSTAC~\cite{plis2016} provides a distributed privacy-preserving computation framework for neuroimaging. 
We apply FL to brain age estimation from structural MRI scans that are distributed across different data silos (cf. Section~\ref{sec:Experiments}).

\paragraph{Privacy.}  Our proposed Semi-Synchronous training protocol can be run under standard privacy techniques such as differential privacy~\cite{abadi2016deep,mcmahan2017learning}, secure multi-party computation (MPC)~\cite{bonawitz2017practical, mohassel2017secureml, kilbertus2018blind}, and homomorphic encryption~\cite{rivest1978data,paillier-crypto,zhang2020batchcrypt,stripelis2021:sipaim}. Currently, we are actively working on a Paillier based additive homomorphic encryption scheme~\cite{paillier1999public} that will encompass all the different training protocols we discuss in this work. We refer the reader to our current progress for additional details~\cite{stripelis2021:sipaim}. Further analysis of privacy techniques is out of scope, since our goal is to evaluate the performance of federated learning training policies in heterogeneous cross-silo settings~\cite{stripelis2021:sipaim}.

\section{Federated Optimization}\label{sec:FederatedOptimization} 

In Federated Learning the goal is to find the optimal set of parameters $w^*$ that minimize the global objective function:
\begin{equation}\label{eq:FederatedFunction}
w^*=\underset{w}{\mathrm{argmin}} f(w) \quad\text{where}\quad f(w)=\sum_{k=1}^{N}\frac{p_k}{\mathcal{P}}F_k(w)
\end{equation}
where N denotes the number of participating learners, $p_k$ the contribution of learner $k$ in the federation,
$\mathcal{P}=\sum p_k$ the normalization factor ($\sum_{k}^N \frac{p_k}{\mathcal{P}}=1$), and $F_k(w)$ the local objective function of learner $k$. 
We refer to the model computed using Equation~\ref{eq:FederatedFunction} as the community model $w_c$.
Every learner computes its local objective by minimizing the empirical risk over its local training set $D_k^T$ as $F_k(w) = \mathbb{E}_{x_k \sim D_k^T}[\ell_k(w;x_k)]$, with $\ell_k$ being the loss function. 
For example, in the FedAvg weighting scheme, the contribution value for any learner $k$ is equal to its local training set size, $p_k = \left|D_k^T\right|$.
The contribution value $p_k$ can be static, or dynamically defined at run time (cf. Section~\ref{sec:FederatedLearningPolicies}).

In the original Federated Learning work~\cite{mcmahan2017communication} every learner aims to minimize its local function $F_k(w)$ using Vanilla Stochastic Gradient Descent (SGD) as its local solver ($\eta=$ learning rate):
\begin{equation}\label{eq:LocalSGD}
    \begin{gathered}
        w_{t+1} = w_{t} - \eta\nabla F_k(w_t)
    \end{gathered}    
\end{equation}

In SGD with Momentum (which can accelerate convergence~\cite{liu2020accelerating}), the local solution $w_{t+1}$ at iteration $t$ is computed as ($u=$ momentum term, $\gamma=$ momentum attenuation factor):
\begin{equation}\label{eq:LocalSGDWithMomentum}
    \begin{gathered}
        u_{t+1} = \gamma u_{t} + \nabla F_k(w_t) \\
        w_{t+1} = w_{t} - \eta u_{t+1}
    \end{gathered}    
\end{equation}

FedProx~\cite{li2020federatedopt} is a variant of the local SGD solver that introduces a proximal term in the update rule to regularize the local updates based on the divergence of the local solution from the global solution (i.e., the community model $w_c$). The local solution $w_{t+1}$ 
is computed as:
\begin{equation}\label{eq:FedproxOptimizer}
    \begin{gathered}
        w_{t+1} = w_{t} - \eta\nabla F_k(w_t) - \eta\mu (w_{t} - w_c)
    \end{gathered}    
\end{equation}
The proximal term $\mu$ controls the divergence of the local solution from the global. In this work, we study the effectiveness of the Semi-Synchronous protocol for all three local SGD solver variants.

\section{Federated Learning Policies}\label{sec:FederatedLearningPolicies}

\begin{figure}[htbp]
    \centering
    \includegraphics[scale=0.5]{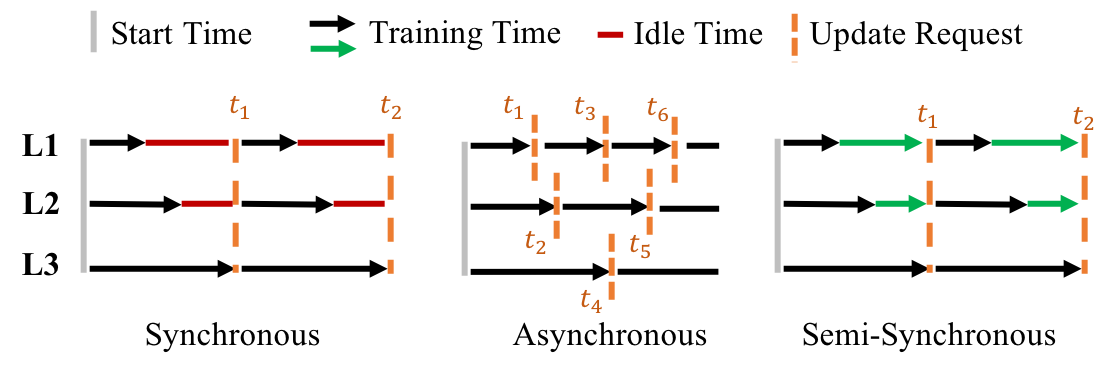}
    \captionsetup{justification=centering}
    \caption{Federated Learning Training Policies: Execution Flow}
    \label{fig:TrainingPolicies}
\end{figure}

In this section we review the main characteristics of synchronous and asynchronous federated learning policies, and introduce our novel semi-synchronous policy. Figure~\ref{fig:TrainingPolicies} sketches their training execution flow. 
We compare these approaches under three evaluation criteria: convergence time, communication cost and energy cost. Rate of convergence is expressed in terms of \textit{parallel processing time}, that is, the time it takes the federation to compute a community model of a given accuracy with all the learners running in parallel. Communication cost is measured in terms of \textit{update requests}, that is, the number of local models sent from any learner to the controller during training. Each learner also receives a community model after each request, so the total number of models exchanged is twice the update requests. Energy cost is based on the \textit{cumulative processing time} (total wall-clock time across learners required to compute the community model) and on the energy efficiency of each learner (e.g., GPU, CPU).
Table~\ref{tbl:TrainingPoliciesCharacteristics} summarizes our findings (cf. Section~\ref{sec:Experiments}).

\begin{table}[tbp]
\begin{tabular}{@{}lccccc@{}}
\toprule
\textbf{Protocol} & processing cost & communication cost & energy cost & idle-free & stale-free \\ \midrule
synchronous       & high                & low            & high & x         & \checkmark          \\
asynchronous      & low               & high             & medium & \checkmark         & x          \\
semi-synchronous  & low                & low             & low & \checkmark         & \checkmark          \\ \bottomrule
\end{tabular}
\caption{Federated Learning Training Policies: Characteristics}
\label{tbl:TrainingPoliciesCharacteristics}
\end{table}

\subsection{Synchronous Federated Learning}
Under a synchronous communication protocol, each learner performs a given number of local steps (usually expressed in terms of local epochs). After all learners have finished their local training, they share their local models with the centralized server (federation controller) and receive a new community model. This training procedure continues for a number of federation rounds (synchronization points). This is a well-established training approach with strong theoretical guarantees and robust convergence for both IID and Non-IID data~\cite{Li2020On,mcmahan2017communication}.

\begin{figure}[htbp]
    \captionsetup[subfigure]{justification=centering}
  \subfloat[2-CNN CIFAR-10\\ (Sync)]{
    \includegraphics[width=0.225\linewidth]{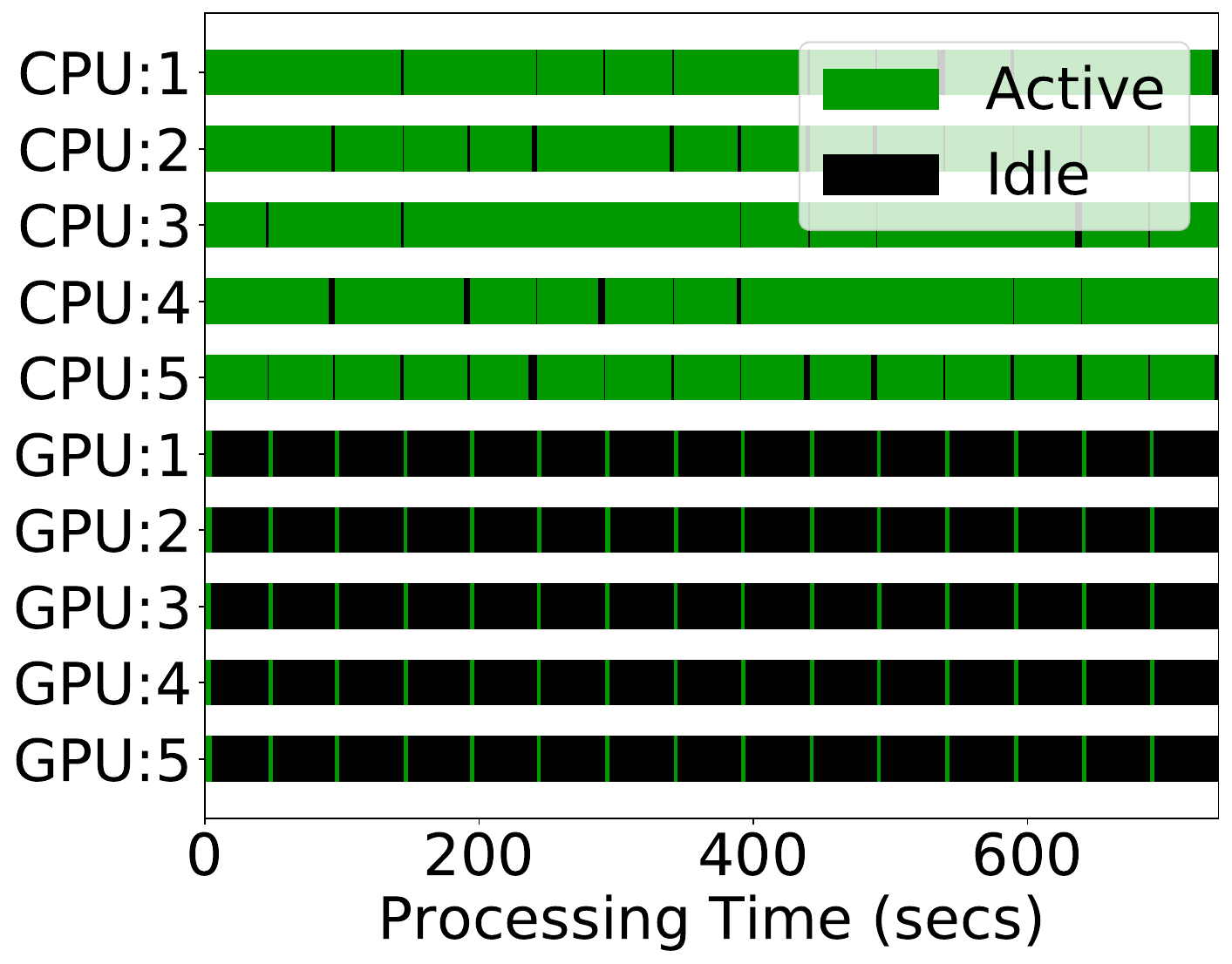}
    \label{subfig:Cifar10_ActiveIdleTime_Sync}
  }
  \subfloat[2-CNN CIFAR-10 \\ (SemiSync)]{
    \includegraphics[width=0.225\linewidth]{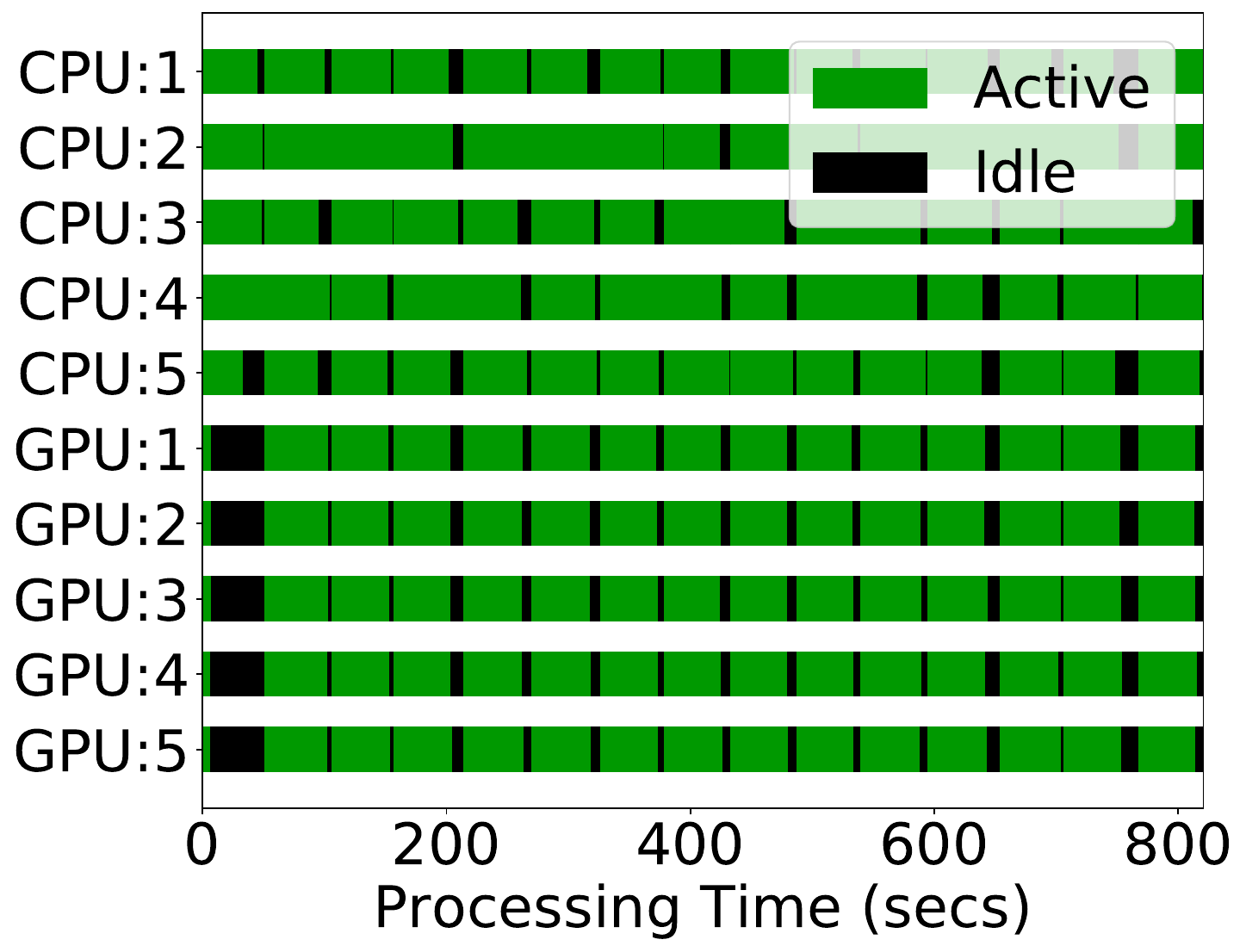}
    \label{subfig:Cifar100_ActiveIdleTime_Sync}
  }
  \subfloat[ResNet-50 CIFAR-100 \\(Sync)]{
    \includegraphics[width=0.225\linewidth]{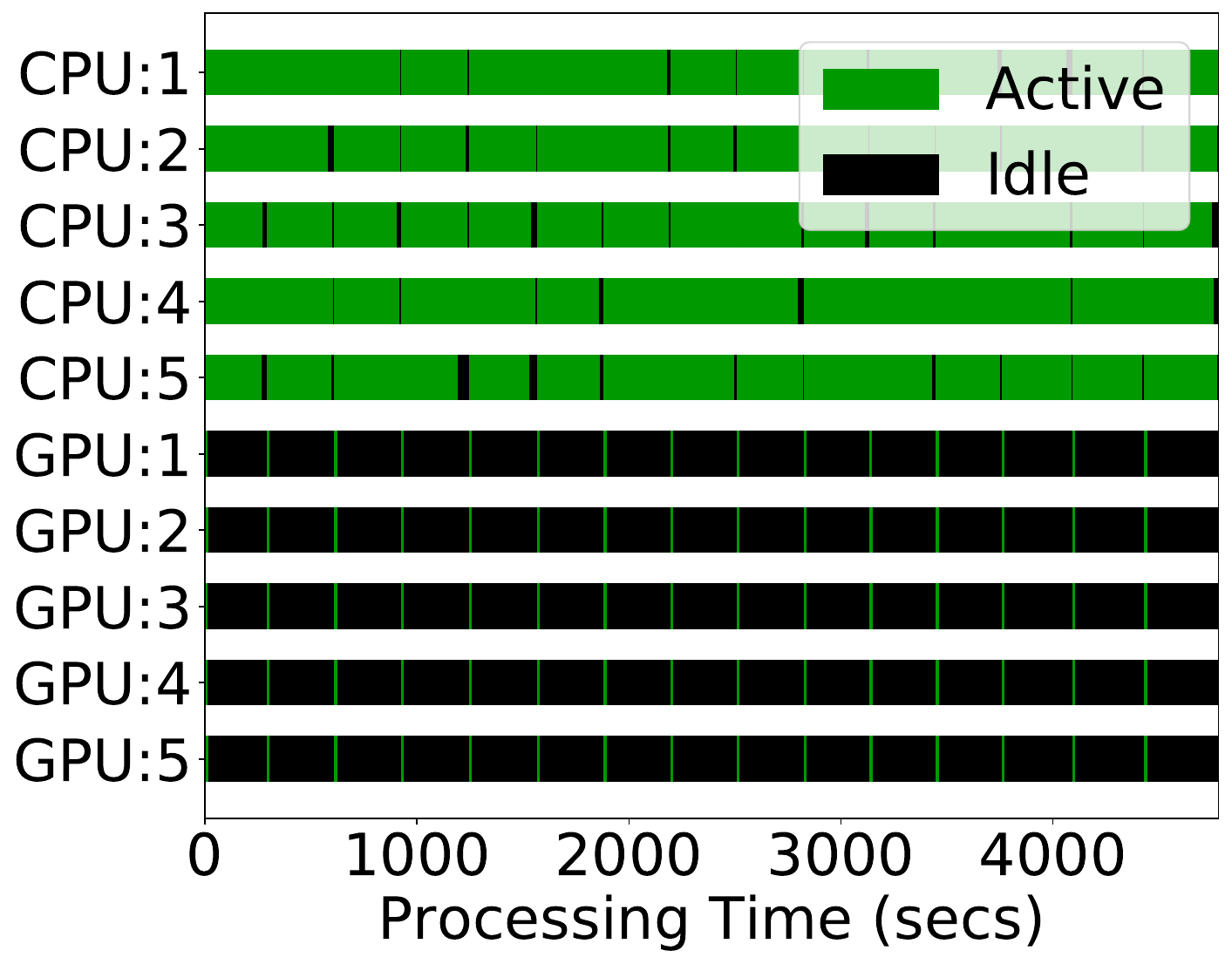}
    \label{subfig:Cifar10_ActiveIdleTime_SemiSync}
  }
  \subfloat[ResNet-50 CIFAR-100 \\ (SemiSync)]{
    \includegraphics[width=0.225\linewidth]{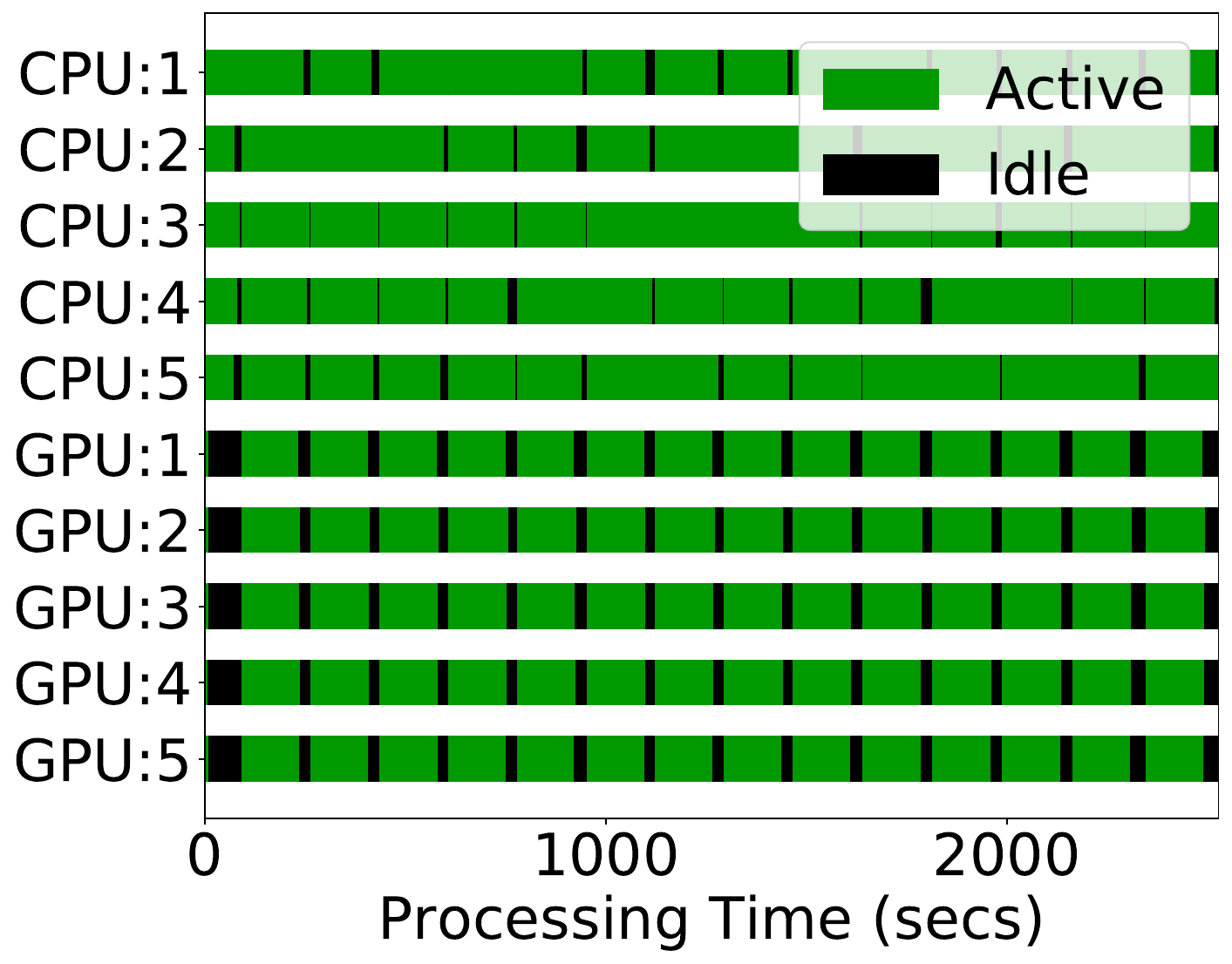}
    \label{subfig:Cifar100_ActiveIdleTime_SemiSync}
  }
  
  \caption{Active vs Idle time for a heterogeneous computational environment with 5 fast (GPUs) and 5 slow (CPUs) learners. Synchronous protocol was run with 4 epochs per learner and Semi-Synchronous with $\lambda=2$ (including cold start federation round).}
  \label{fig:ActiveVSIdleTime}
\end{figure}

However, a limitation of synchronous policies is their slow convergence due to waiting for slow learners (stragglers~\cite{dean2013tail}). For a federation of learners with heterogeneous computational capabilities, fast learners remain idle most of the time, since they need to wait for the slow learners to complete their local training before a new community model can be computed (Figure~\ref{fig:TrainingPolicies}). As we move towards larger networks, this resource underutilization is exacerbated. 
Figure~\ref{fig:ActiveVSIdleTime}(a,b) shows idle times of a synchronous protocol when training a 2-CNN (a) or a ResNet-50 (b) network in a federation with fast (GPU) and slow (CPU) learners. The fast learners are severely underutilized.

\subsection{Asynchronous Federated Learning}
In asynchronous Federated Learning no synchronization point exists and learners can request a community update from the controller whenever they complete their local assigned training. Asynchronous protocols have faster convergence speed since no idle time occurs for any of the participating learners. However, they incur higher communication cost and lower generalizability due to staleness~\cite{cui2014exploiting,dai2018toward}.    
Figure \ref{fig:TrainingPolicies}(center) illustrates a typical asynchronous policy. The timestamps $t_i$ represent the update requests issued by the learners to the federation controller. No synchronization exists and every learner issues an update request at its own learning pace. 
All learners run continuously, so there is no idle time. 

Since in asynchronous protocols, no strict consistency model~\cite{lamport1979make} exists, it is inevitable for learners to train on \textit{stale} models. Community model updates are not directly visible to all learners and different staleness degrees may be observed~\cite{ho2013more,cui2014exploiting}. Recently, \textit{FedAsync}~\cite{xie2019asynchronous} was proposed as an asynchronous training policy for Federated Learning by weighting every learner in the federation based on functions of model staleness. FedAsync defines staleness as the difference between the current global timestamp (vector clock~\cite{mattern1988virtual}) of the community model and the global timestamp associated with the committing model of a requesting learner. Specifically, for learner $k$, its staleness value is equal to $\mathcal{S}_k= (T - \tau_k + 1)^{-1/2}$, (i.e., FedAsync $+$ Poly) with $T$ being the current global clock value and $\tau_k$ the global clock value of the committing model of the requesting learner. 

Given that staleness can also be controlled by tracking the total number of iterations or number of steps (i.e., batches) applied on the community model, we propose a new asynchronous protocol, \textit{FedRec}, which weights models based on \textit{recency}, extending the notion of \textit{effective staleness} in~\cite{dai2018toward,dai2018learning}. For each model we define the number of steps $\mathtt{s}$ that were performed in its computation. Assume a learner $k$ that receives a community model $c^t$ at time $t$, which was computed over a cumulative number of steps $\mathtt{s}_c^{t}$ (the sum of steps used by each of the local models involved in computing the community model). Learner $k$ then performs $\mathtt{s}_k$ local steps starting from this community model, and requests a community update at time $t^{\prime}$. By that time the current community model may contain $\mathtt{s}_c^{t^{\prime}} (>\mathtt{s}_c^{t}) $ steps, since other learners may have contributed steps between $t$ and $t^{\prime}$. Therefore, the effective staleness weighting value of learner $k$ in terms of steps is $\mathcal{S}_k=(\mathtt{s}_c^{t^{\prime}} - ( \mathtt{s}_c^{t} + \mathtt{s}_k))^{-1/2}$ (following the FedAsync + Poly function). When $\mathtt{s}_c^{t^{\prime}} - ( \mathtt{s}_c^{t} + \mathtt{s}_k) < 0$, $\mathcal{S}_k$ is set to 1. 
As shown in Section~\ref{sec:Experiments}, the step-based recency/staleness function of FedRec outperforms the time-based staleness function of FedAsync (cf. Figures \ref{fig:Cifar10HeterogeneousCluster}, \ref{fig:Cifar100HeterogeneousCluster}, \ref{fig:EMNISTHeterogeneousCluster}).

\subsection{Semi-Synchronous Federated Learning}
We have developed a novel \textit{Semi-Synchronous} training policy that seeks to balance resource utilization and communication costs%
\footnote{In our cross-silo settings, we do not consider delays due to learners' transmission speed. In our experiments in Section~\ref{sec:Experiments} model transmission time is less than 0.5\% of the computation time.}
(cf. Figure~\ref{fig:ActiveVSIdleTime}). In this policy every learner continues training up to a specific synchronization time point (cf. Figure~\ref{fig:TrainingPolicies}(right)). The synchronization point is based on the maximum time it takes for any learner to perform a single epoch ($t_{k}^{e}$). Specifically:
\begin{equation}\label{eq:SemiSynchronousScheduling}
    \begin{gathered}
        t_{k}^{e} = \frac{|D_k^T|}{\beta_k} t_{\beta_k}, \;\; \beta_k,t_{\beta_k} > 0; 
        \quad
        t_{max}(\lambda) = \lambda \max \limits_{k \in N} {\{ t_{k}^{e}  \}}, \;\; \lambda > 0;
        \quad   
        \mathcal{B}_k = \dfrac{t_{max}}{t_{\beta_{k}}}, \;\; \forall k \in N 
    \end{gathered}    
\end{equation}
where $D_k^T$ refers to the local training data size of learner $k$, $\beta_k$ to the batch size of learner $k$ and $t_{\beta_k}$ to the time it takes learner $k$ to perform a single step (i.e., process a single batch). The hyperparameter $\lambda$ controls the number of local passes the slowest learner in the federation needs to perform before all learners synchronize. For example, $\lambda=2$ refer to the slowest learner completing two epochs. The hyperparmeter $\lambda$ can be fractional, that is, the slowest learner may only process part of its training set in a federation round. The term $\mathcal{B}_k$ denotes the number of steps (batches) learner $k$ needs to perform before issuing an update request, which depends on its computational speed. A theoretical analysis of the weight divergence of our proposed Semi-Synchronous training scheme compared to the centralized model and a more detailed formulation of the federated optimization problem can be found in the Appendix.

To compute the necessary statistics (i.e. time-per-batch per learner), \textit{SemiSync} performs an initial cold start federation round (see GPUs in  Figure~\ref{fig:ActiveVSIdleTime}(b,d)) where every learner trains for a single epoch and the controller collects the statistics to synchronize the new SemiSync federation round. Here, the hyperparameter $\lambda$ and the timings per batch are kept static throughout the federation training once  defined, although others schedules are possible (cf. Section~\ref{sec:Discussion}). 
To obtain a good estimate of the processing time per batch, in the cold start phase the system has  every learner complete a full epoch. The system sets a maximum duration for cold start to prevent a very slow learner from disrupting the federation.

In our SemiSync approach, the learners' synchronization point does not depend on the number of completed epochs, but on the synchronization period. Learners with different computational power and amounts of data perform a different number of epochs, including fractional epochs. There is no idle time. Since the basic unit of computation is the batch, this allows for a more fine-grained control on when a learner contributes to the community model. This policy is particularly beneficial in heterogeneous computational and data distribution environments. 

\subsection{Training Policies Cost Analysis} \label{subsec:TrainingPoliciesCostAnalysis}

\paragraph{Parallel Processing Time} 
We are interested in the wall-clock time it takes the federation to reach a community model of a given accuracy, with all learners running in parallel. For synchronous and semi-synchronous protocols, this is simply the number of federation rounds ($R$) times the synchronization period ($t_{max}(\lambda)$)~\cite{luo2020cost}.  
For asynchronous protocols, this is the time at which a learner submits the last local model that makes the community model reach the desired accuracy (e.g., time $t_6$ in Figure \ref{fig:TrainingPolicies}). If $pt$ denotes the processing time difference between update requests, and $U$ the number of update requests, then the total parallel processing time is computed as: 
\begin{equation}\label{eq:ProcessingCostFunction}
    PT_{Sync/SemiSync}= R \, t_{max}(\lambda);
    \quad
    PT_{Async} = \sum_{i=1}^U pt_i
\end{equation}

\paragraph{Communication Cost}
We measure communication cost by the total number of update requests issued during training by the learners to the federation controller. Each update request accounts for two model exchanges: the learner sends its local model to the controller and receives the community model. In a federation of $N$ learners, for synchronous and semi-synchronous protocols with $R$ synchronization points, and for asynchronous with $U$ update requests: 
\begin{equation}\label{eq:CommunicationCostFunction}
    CC_{Sync/SemiSync} = N \, R    \qquad CC_{async} = U
\end{equation}

\paragraph{Energy Cost} 
The energy cost is based on the \textit{cumulative processing time} of all learners to complete their local training~\cite{luo2020cost} weighted by the energy cost ($\epsilon_k$) of each learner's processor (e.g., GPU or CPU). For asynchronous protocols, let $\Lambda_k$ denote the total number of local epochs performed by a learner k to reach a particular timestamp, then the cumulative energy cost is: 

\begin{equation}\label{eq:EnergyCostFunction}
    EC_{Sync} = R \, \lambda \, \sum_{k=1}^N \epsilon_k \, t_{k}^{e} 
    \qquad 
    EC_{SemiSync} = R \, t_{max}(\lambda) \sum_{k=1}^N \epsilon_k 
    \qquad 
    EC_{Async} = \sum_{k=1}^N \epsilon_k \, \Lambda_k \, t_{k}^{e}
\end{equation}

\section{Federated Learning Environment}
\label{sec:MetisFederatedLearningFramework}

\begin{figure}[htpb]
\begin{minipage}{.55\textwidth}
  \hspace*{-0.5cm}
  \includegraphics[width=\linewidth]{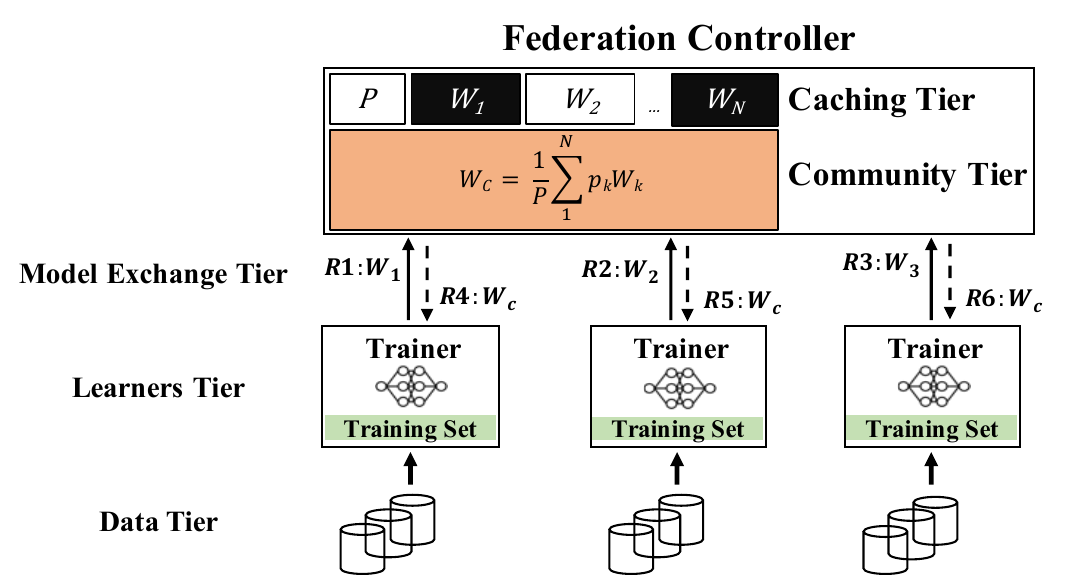}
  \captionof{figure}{Metis System Architecture}
  \label{fig:MetisSystemArchitecture}
\end{minipage}%
\begin{minipage}{.4\textwidth}
  \includegraphics[width=\linewidth]{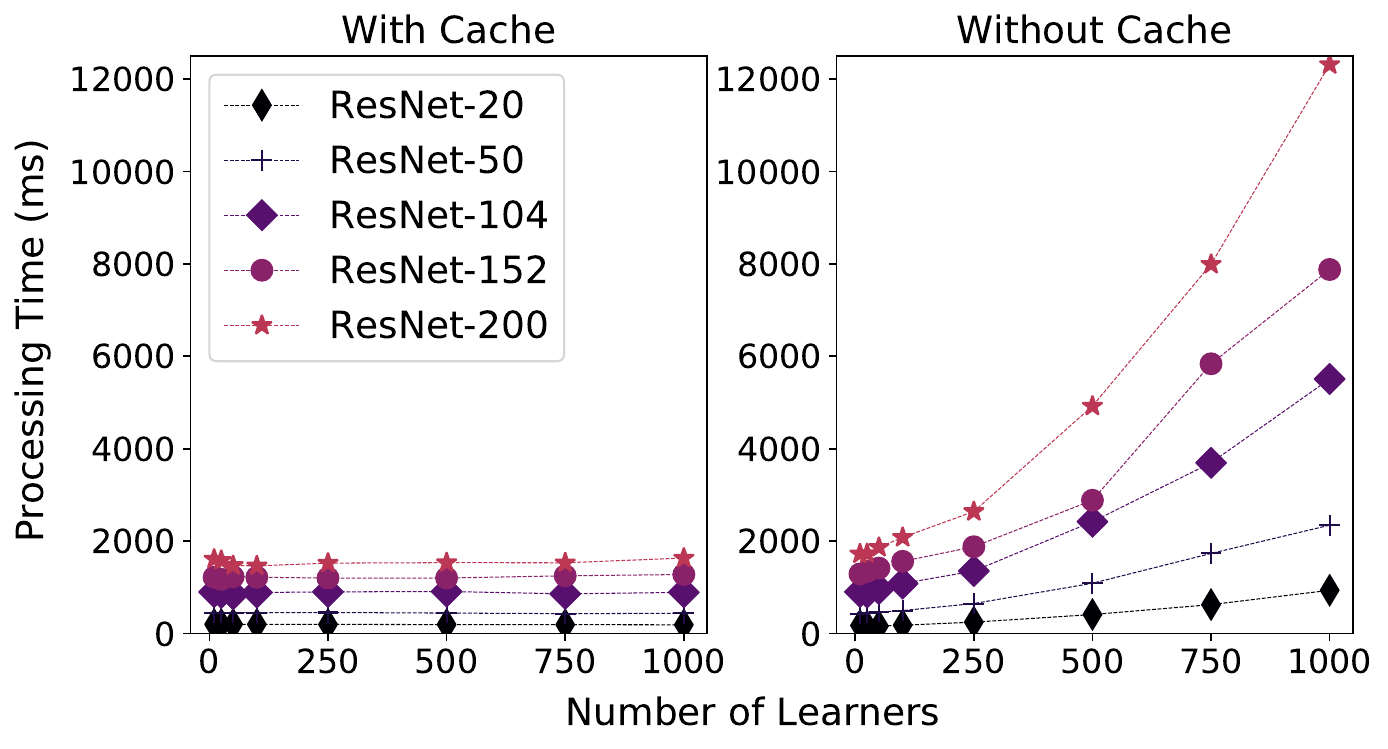}
  \captionof{figure}{Community model computation with (left) and without (right) caching.}
  \label{fig:CommunityWithAndWithoutCache}
\end{minipage}
\end{figure}

We have designed and developed a flexible Federated Learning system, called Metis, to explore  different communication protocols and model aggregation weighting schemes (Figure~\ref{fig:MetisSystemArchitecture}). Metis uses Tensorflow~\cite{abadi2016tensorflow} as its deep learning execution engine. In Figure~\ref{fig:MetisSystemArchitecture}, we decompose the federated learning environment into tiers in order to exemplify the data management and training procedures occurring in cross-silo settings.

\paragraph{Federation Controller} The centralized controller is a multi-threaded process with a modular design that integrates a collection of extensible microservices (e.g., caching and community tiers). The controller orchestrates the execution of the entire federation and is responsible to initiate the system pipeline, broadcast the initial community model and handle learners' update requests. Every incoming update request is handled by the controller in a FIFO ordering through a mutual exclusive lock, ensuring system state linearizability~\cite{herlihy1990linearizability}. Essentially, the federation controller is a materialized version of the Parameter Server~\cite{abadi2016tensorflow,dean2012large} concept, widely used in distributed learning applications. The centralized (star-shaped~\cite{bellavista2021decentralised}) federated learning topology that we investigate in this work, i.e., a centralized coordinator and a collection of learning nodes, is considered a standard topology in cross-silo federated learning settings~\cite{li2020federated,yang2019federated,kairouz2019advances}. Compared to existing work~\cite{bonawitz2019towards,wang2020industrial,gao2020salient} that follows a hierarchical structure, where multiple coordinators may exist along with a set of master/cloud-networks and sub-aggregators/fog-networks, our learning environment consists of a single master coordinator (i.e., Federation Controller) that is responsible for coordinating the federated execution and aggregating the learners’ local models.

\paragraph{Community \& Caching Tier.} The community tier computes a new community model $w_c$ as a weighted average of the most recent model that each learner has shared with the controller (Eq.~\ref{eq:FederatedFunction}). To facilitate this computation, it is natural to store the most recently received local model of every learner in-memory or on disk. Therefore, the memory and storage requirements of a community model depend on the number of learners contributing to the community model.
Similarly to the computation of the community model in synchronous protocols, we extend this approach to asynchronous protocols by using our proposed caching scheme at every update request.
For synchronous protocols we always need to perform a pass over the entire collection of stored local models, with a computational cost $O(MN)$, where $M$ is the size of the model and $N$ is the number of participating learners.
For asynchronous protocols where learners generate update requests at different paces and the total number of update requests is far greater than synchronous, such a repetitive complete pass is expensive and we can leverage the existing cached/stored local models to compute a new community model in $O(M)$ time, independent of the number of learners.

Consider an unnormalized community model consisting of $m$ matrices, $w_c=\langle w_{c_1}, w_{c_2}, \ldots, w_{c_m}\rangle$, and a community normalization weighting factor $\mathcal{P}=\sum_{k=1}^{N} p_{k}$. 
Given a new request from learner $k$, with new community contribution value $p_k^{\prime}$, the new normalization value $\mathcal{P^{\prime}}$ is  $\mathcal{P^{\prime}} = \mathcal{P} + p_k^{\prime} - p_k$, where $p_k$ is the learner's previous contribution value.
For every component matrix $w_{c_i}$ of the community model, the updated matrix $w_{c_i}^{\prime}$ is  $w_{c_i}^{\prime} = w_{c_i} + p_k^{\prime} w_{k,i}^{\prime} - p_k w_{k,i}$, where $w_{k,i}^{\prime}, w_{k,i}$ are the new and existing component matrices of learner $k$. The new community model is $w_c = \frac{1}{\mathcal{P}^{\prime}} w_c^{\prime}$. Using this caching approach, in asynchronous execution environments, the most recently contributed local model of every learner in the federation is always considered in the community model.

Some existing asynchronous community mixing approaches~\cite{xie2019asynchronous,sprague2018asynchronous} compute a weighted average using a mixing hyperparameter between the current community and the committing local model of a requesting learner. In contrast, our caching approach eliminates this mixing hyperparameter dependence and performs a weighted aggregation using all recently contributed local models. Figure~\ref{fig:CommunityWithAndWithoutCache} shows the computation cost for different sizes of a ResNet community model (from Resnet-20 to Resnet-200), in the CIFAR-100 domain, as the federation increases to 1000 learners. With our caching mechanism the time to compute a community model remains constant irrespective of the number of participating learners, while it significantly increases without it. 

\paragraph{Model Exchange Tier.} This tier follows a stateful client-server communication architecture~\cite{hauswirth1999component}. Every learner in the federation contacts the controller for a community update once it completes its locally assigned training and shares its local model with the controller. Upon computing the new community model, in the synchronous and semi-synchronous cases, the controller broadcasts the new model to all learners, while in the asynchronous case, the controller sends the new model to the requesting learner. Figure~\ref{fig:MetisSystemArchitecture} represents the model exchange of the semi/synchronous cases with requests R1 to R6.

\paragraph{Learners \& Data Tier.} In the current work, all learners train on the same neural network architecture with identical hyperparameter values (learning rate, batch size, etc.), starting from the same initial (random) model state, and using the same local SGD optimizer (but see future work in Section~\ref{sec:Discussion}). 
The number of local steps a learner performs before issuing an update request are defined either in terms of epochs/batches or the maximum scheduled time (see Eq. \ref{eq:SemiSynchronousScheduling}). Every learner trains on its own local training dataset and no data is shared.

\paragraph{Execution Pipeline.} Algorithm \ref{alg:METIS} describes the execution pipeline within Metis for synchronous, semi-synchronous and asynchronous communication protocols. In synchronous and semi-synchronous protocols, the controller waits for all the participating learners to finish their local training task before it computes a community model, distribute it to the learners, and proceed to the next global iteration.
In asynchronous protocols, the controller computes a community model whenever a single learner finishes its local training, using the caching mechanism, and sends the new model to the learner. 
In all cases, the controller assigns a contribution value $p_k$ to the local model $w_k$ that a learner $k$ shares with the community. For synchronous and asynchronous FedAvg (SyncFedAvg and AsyncFedAvg), this value is statically defined and based on the size of the learner's local training dataset, $\left|D_k^T\right|$. For other weighting schemes, such as FedRec, the \textsc{Staleness} procedure computes it dynamically. 
The \textsc{LearnerOpt} procedure implements the local training of each learner. The local training task assignment information is passed to every learner through the metadata, $meta$, collection. Finally, a learner trains on its local model using either Vanilla SGD, Momentum SGD or FedProx as its local SGD solver.

\begin{algorithm*}[tpb]
    \caption{\texttt{FL Training with Metis.} Community model $w_c$, comprising $m$ matrices, is computed from $N$ learners; $\gamma =$ momentum attenuation factor; $\eta =$ learning rate; $\beta = $ batch size.}
    \label{alg:METIS}
    \vspace{-\baselineskip}
    \begin{multicols}{2}
        \begin{algorithmic}
        \renewcommand{\algorithmicrequire}{\textbf{Initialization}: $w_c, \gamma, \eta, \beta$}
        \REQUIRE
        \STATE
        \renewcommand{\algorithmicrequire}{\textbf{\underline{(Semi-)Synchronous}}}
        \REQUIRE
        \FOR{$t = 0, \dots, T-1$}
             \FOR{each learner $k \in N$ \textbf{in parallel}}
                \STATE $w_k = \textsc{LearnerOpt}(w_c, meta)$
                \STATE $p_k = \left|D_k^T\right| \halfquad \text{(SyncFedAvg)}$
             \ENDFOR
             \STATE {$w_c = \sum_{k=1}^{N}\frac{p_{k}}{\mathcal{P}}w_k$ with $\mathcal{P}=\sum_{k}^N p_k$}
             \STATE Reply $w_c$ to every learner
        \ENDFOR
        \STATE
        \renewcommand{\algorithmicrequire}{\textbf{\underline{Asynchronous}}}
        \REQUIRE
        \STATE $P=0$; $\forall k \in N, p_k=0$; $\forall i \in m, W_{c,i}=0$
        \STATE $\forall k \in N$ \textsc{LearnerOpt}($w_c, meta$)
        \WHILE{\textbf{true}}
            \IF{(learner $k$ requests update)}
                \STATE $p_k^{\prime} = 
                    \begin{cases} 
                        \left|D_k^T\right| \halfquad \text{\small (AsyncFedAvg)} \\ 
                        \textsc{Staleness}(k) \halfquad \text{\small (FedRec)} \\ 
                    \end{cases}$
                \STATE $\mathcal{P}^{\prime} = \mathcal{P} + p_k^{\prime} - p_k$
                \FOR{$i \in m$}
                    \STATE $W_{c,i}^{\prime} = W_{c,i} + p_k^{\prime} w_{k,i}^{\prime} - p_k w_{k,i}$
                \ENDFOR    
                \STATE $w_c^{\prime} = \frac{1}{\mathcal{P}^{\prime}} W_c^{\prime}$
                \STATE Reply $w_c^{\prime}$ to learner $k$
            \ENDIF
        \ENDWHILE
        \columnbreak
        \renewcommand{\algorithmicrequire}{\textbf{\textsc{LearnerOpt($w_t, meta$):}}}
        \REQUIRE        
            \STATE $\mathcal{B} = 
            \begin{cases} 
                \text{meta[epochs]} * \sfrac{D_k^{T}}{\beta} \halfquad \text{\small (Sync \& Async)} \\
                \text{meta[}t_{max}\text{]} / {t_{\beta_{k}}} \halfquad \text{\small (SemiSync, cf. Section \ref{sec:FederatedLearningPolicies})} \\ 
            \end{cases}$
            \STATE $\mathcal{B} =$ Shuffle $\mathcal{B}$ training batches of size $\beta$
            \FOR{$b \in \mathcal{B}$}
                \IF{Vanilla SGD}
                    \STATE {$w_{t+1} = w_{t} - \eta\nabla F_k(w_t;b)$}
                \ENDIF    
                \IF{Momentum}
                    \STATE {$u_{t+1} = \gamma u_{t} - \eta\nabla F_k(w_t;b)$}
                    \STATE {$w_{t+1} = w_{t} + u_{t+1}$}
                \ENDIF
                \IF{FedProx}
                    \STATE {$w_{t+1} = w_{t} - \eta\nabla F_k(w_t;b) - \eta\mu (w_{t} - w_c)$}
                \ENDIF
            \ENDFOR
            \STATE Reply $w_{t+1}$ to controller
        \STATE
        \renewcommand{\algorithmicrequire}{\textbf{\textsc{Staleness($k$):}}}
        \REQUIRE
        \STATE $\mathtt{s}_c^{t^{\prime}}=$ current time community model steps
        \STATE $\mathtt{s}_c^{t}=$ previous time community model steps
        \STATE $\mathtt{s}_k=$ learner $k$ steps between $t$ and $t^\prime$
        \STATE $\delta = \mathtt{s}_c^{t^{\prime}} - (\mathtt{s}_c^{t} + \mathtt{s}_k)$
        \STATE $\mathcal{S}_k =
        \begin{cases} 
            \delta^{-1/2}, \delta>0 \\
            1, \delta\leq0
            \end{cases}$
        \STATE Reply $\mathcal{S}_k$
    \end{algorithmic}
    
    \end{multicols}
    \vspace{-2mm}
\end{algorithm*}

\section{Experiments} \label{sec:Experiments}
We conduct an extensive experimental evaluation of different training policies on a diverse set of federated learning environments with heterogeneous amounts of data per learner, local data distributions, and computational resources. We evaluate the protocols on the CIFAR-10, CIFAR-100 and ExtendedMNIST By Class~\cite{cohen2017emnist,caldas2018leaf} benchmark datasets with a federation consisting of 10 learners as well as on the BrainAge prediction task~\cite{cole2017predicting,jonsson2019brain,PENG2020101871,stripelis2021scaling,gupta2021improved} with a federation of 8 learners. 
The asynchronous protocols (i.e., FedRec and AsyncFedAvg) were run using the caching mechanism described in Section~\ref{sec:MetisFederatedLearningFramework} except for FedAsync. FedAsync was run using the polynomial staleness function, i.e., FedAsync$+$Poly, with mixing hyperparameter $a=0.5$ and model divergence regularization factor $\rho=0.005$, which is reported to have the best performance~\cite{xie2019asynchronous}.

\paragraph{Models Architecture.} 
The architecture of the deep learning networks for CIFAR-10 and CIFAR-100 come from the Tensorflow tutorials: for CIFAR-10 we train a 2-CNN%
\footnote{CIFAR-10: \url{https://github.com/tensorflow/models/tree/r1.13.0/tutorials/image/cifar10}} 
and for CIFAR-100 a ResNet-50%
\footnote{CIFAR-100: \url{https://github.com/tensorflow/models/tree/r1.13.0/official/resnet}}.
The 2-CNN model%
\footnote{ExtendedMNIST: \url{https://github.com/TalwalkarLab/leaf/blob/master/models/femnist/cnn.py}}
architecture for ExtendedMNIST comes from the LEAF benchmark~\cite{caldas2018leaf}.
The 5-CNN for BrainAge is from~\cite{stripelis2021scaling}. For all models, during training, we share all trainable weights (i.e., kernels and biases). For ResNet we also share the batch normalization, gamma and beta matrices. The random seed for all our experiments is set to 1990.

\paragraph{Models Hyperparameters.} For CIFAR-10 in homogeneous and heterogeneous environments the synchronous and semi-synchronous protocols were run with Vanilla SGD, Momentum SGD and FedProx;  asynchronous protocols (FedRec, AsyncFedAvg) were run with Momentum SGD. 
For CIFAR-100 all the methods were run with Momentum (following the tutorial recommendation). 
For ExtendedMNIST By Class (following the benchmark recommendation) and BrainAge, we used Vanilla SGD. 
We originally performed a grid search, on the centralized model, over different combinations of learning rate $\eta$, momentum factor $\gamma$, and mini batch size $\beta$. For the proximal term $\mu$ in FedProx, we used the values from the original work~\cite{li2020federatedopt}. After identifying the optimal combination, we kept the hyperparameter values fixed throughout the federation training. In particular, for CIFAR-10 we used $\eta$=0.05, $\gamma$=0.75, $\mu=0.001$ and $\beta$=100, for CIFAR-100, $\eta$=0.1, $\gamma$=0.9, $\beta$=100, for ExtendedMNIST, $\eta$=0.01 and $\beta$=100, and for BrainAge $\eta$=$5\text{x}10^{-5}$ and $\beta$=1. For both synchronous and asynchronous policies, we originally evaluated the convergence rate of the federation under different numbers of local epochs $\{1,2,4,8,16,32\}$ and we observed the best performance when assigning $4$ local epochs per learner. For the semi-synchronous case, we investigated the convergence of hyperparameter $\lambda$ within the set $\{0.5,1,2,3,4\}$.

\paragraph{Computational Environment.} Our homogeneous federation environment for CIFAR and ExtendedMNIST consists of 10 fast learners (GPUs) and our heterogeneous of 5 fast (GPU) and 5 slow (CPU) learners. For the BrainAge task our homogeneous environment consists of 8 fast learners. The fast learners were run on a dedicated GPU server equipped with 8~GeForce GTX 1080 Ti graphics cards of 10~GB RAM each, 40~Intel(R) Xeon(R) CPU E5-2630 v4 @ 2.20GHz, and 128GB DDR4 RAM. The slow learners were run on a separate server equipped with 48 Intel(R) Xeon(R) CPU E5-2650 v4 @ 2.20GHz and 128GB DDR4 RAM. For the \mbox{2-CNN} used in CIFAR-10 the processing time per batch for fast learners is: $t_{\beta_{k}} \approx 30ms$, and for slow: $t_{\beta_{k}} \approx 300ms$, for the ResNet-50 used in CIFAR-100, for fast is: $t_{\beta_{k}} \approx 60ms$ and for slow: $t_{\beta_{k}} \approx 2000ms$, for the \mbox{2-CNN} used in ExtendedMNIST for fast is: $t_{\beta_{k}} \approx 50ms$ and for slow: $t_{\beta_{k}} \approx 800ms$ and for the \mbox{5-CNN} used in BrainAge for fast is: $t_{\beta_{k}} \approx 120ms$.

\paragraph{Data Distributions.} We evaluate the training policies over multiple environments with heterogeneous data sizes, data distributions, and learning problems (classification for CIFAR \& ExtendedMNIST, regression for BrainAge).%
\footnote{CIFAR \& ExtendedMNIST: \url{https://dataverse.harvard.edu/dataset.xhtml?persistentId=doi\%3A10.7910\%2FDVN\%2FPQ34F8}
BrainAge: \url{https://dataverse.harvard.edu/dataset.xhtml?persistentId=doi:10.7910/DVN/2RKAQP}
}
We consider three types of \textit{data size distributions}: \textit{Uniform}, where every learner has the same number of examples; \textit{Skewed}, where each learner has a progressively smaller set of examples; and \textit{Power Law} (exponent=1.5) to model extreme variations in the amount of data (intended to model the long tail of science). Learners train on all available local examples.

To model \textit{statistical heterogeneity} in classification tasks, we assign a different number of examples per class per learner. 
Specifically, with \textit{IID} we denote the case where all learners hold training examples from all the target classes, and with \textit{Non-IID(x)} we denote the case where every learner holds training examples from only x classes. For example, Non-IID(3) in CIFAR-10 means that each learner only has training examples from 3 target classes (out of the 10 classes in CIFAR-10). 
For Power Law data sizes and Non-IID configurations, in order to preserve scale invariance, we needed to assign data from more classes to the learners at the head of the distribution. For example, for CIFAR-10 with Power Law and a goal of 5 classes per learner, the actual distribution is Non-IID(8x1,7x1,6x1,5x7), meaning that the first learner holds data from 8 classes, the second from 7 classes, the third from 6 classes, and all 7 subsequent learners hold data from 5 classes. For brevity, we refer to this distribution as Non-IID(5). Similarly for CIFAR-10 Power Law and Non-IID(3), the actual distribution is Non-IID(8x1,4x1,3x8). For CIFAR-100, Power Law and Non-IID(50), the actual distribution is Non-IID(84x1,76x1,68x1,64x1,55x1,50x5). 

\begin{figure*}[tbp]
  \captionsetup[subfigure]{justification=centering}
  \centering
  \subfloat[CIFAR-10 \\ Uniform \& Non-IID(5)]{
    \centering\includegraphics[width=0.3\linewidth]{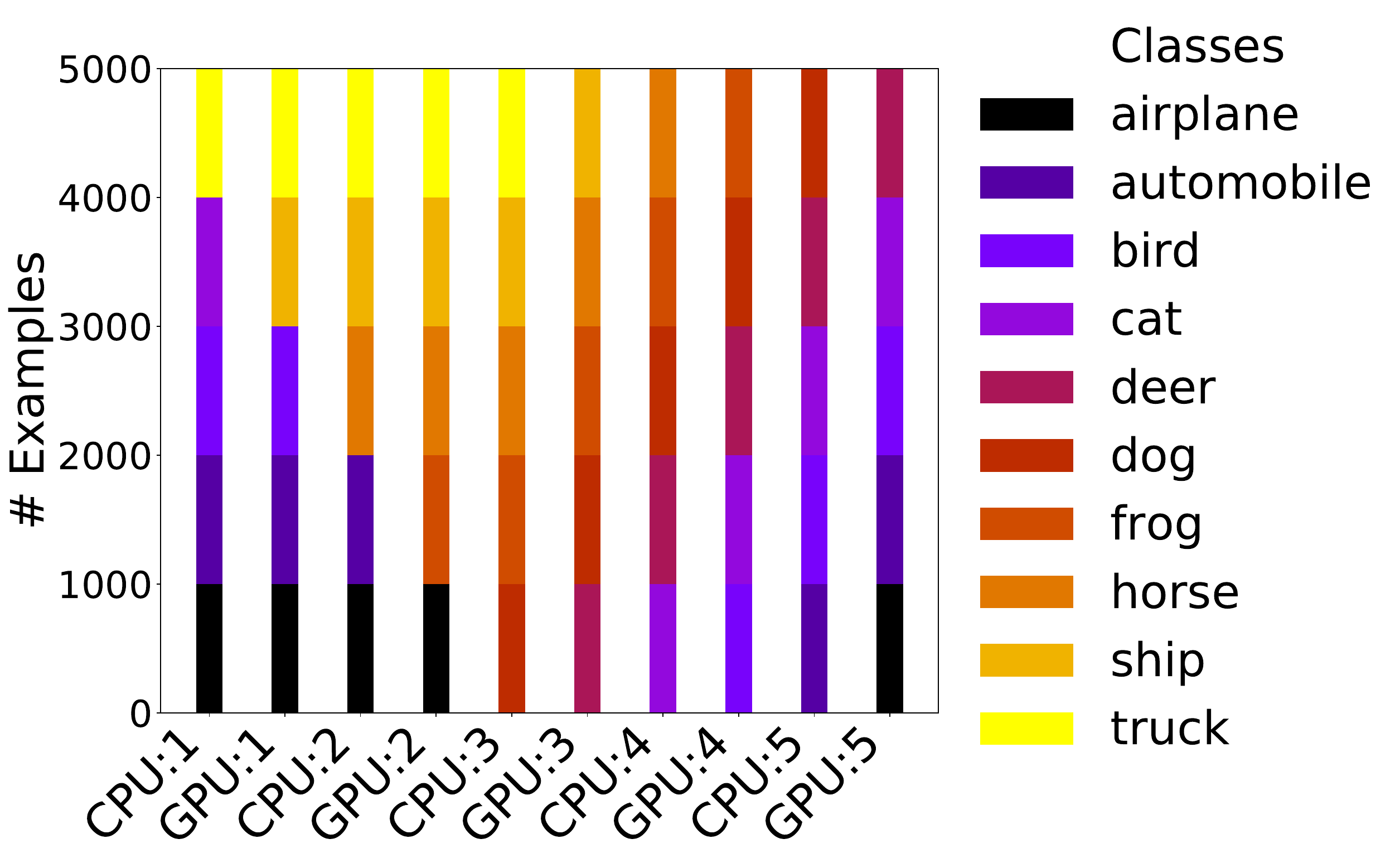}
    \label{subfig:Cifar10_Uniform_NonIID5_DataDistribution}
  }
  \subfloat[CIFAR-100 \\ Skewed \& Non-IID(50)]{
    \centering\includegraphics[width=0.3\linewidth]{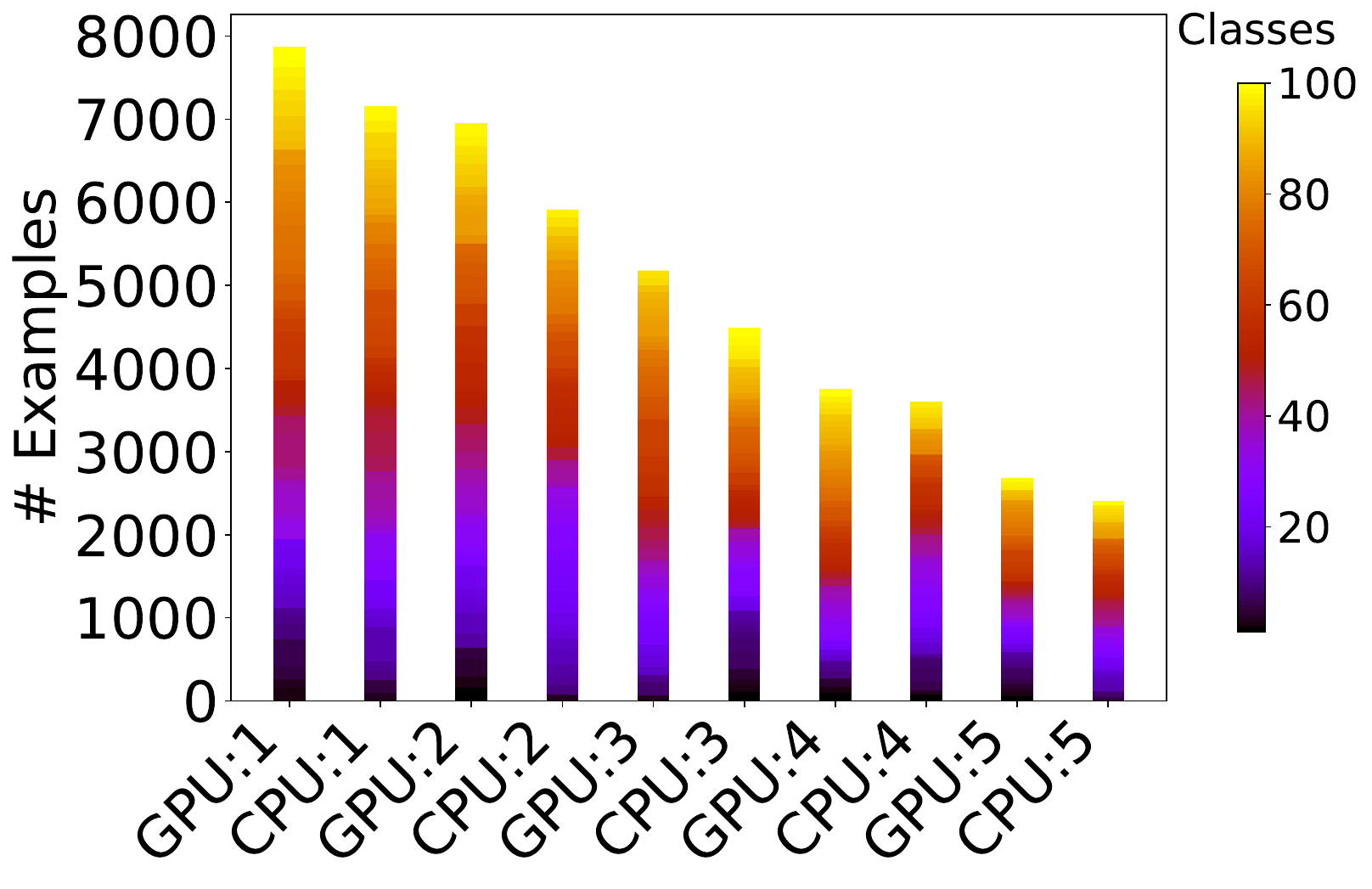}
    \label{subfig:Cifar100_Skewed_NonIID50_DataDistribution}
  }
  \subfloat[ExtendedMNIST By Class \\ Power Law \& IID]{
    \centering\includegraphics[width=0.3\linewidth]{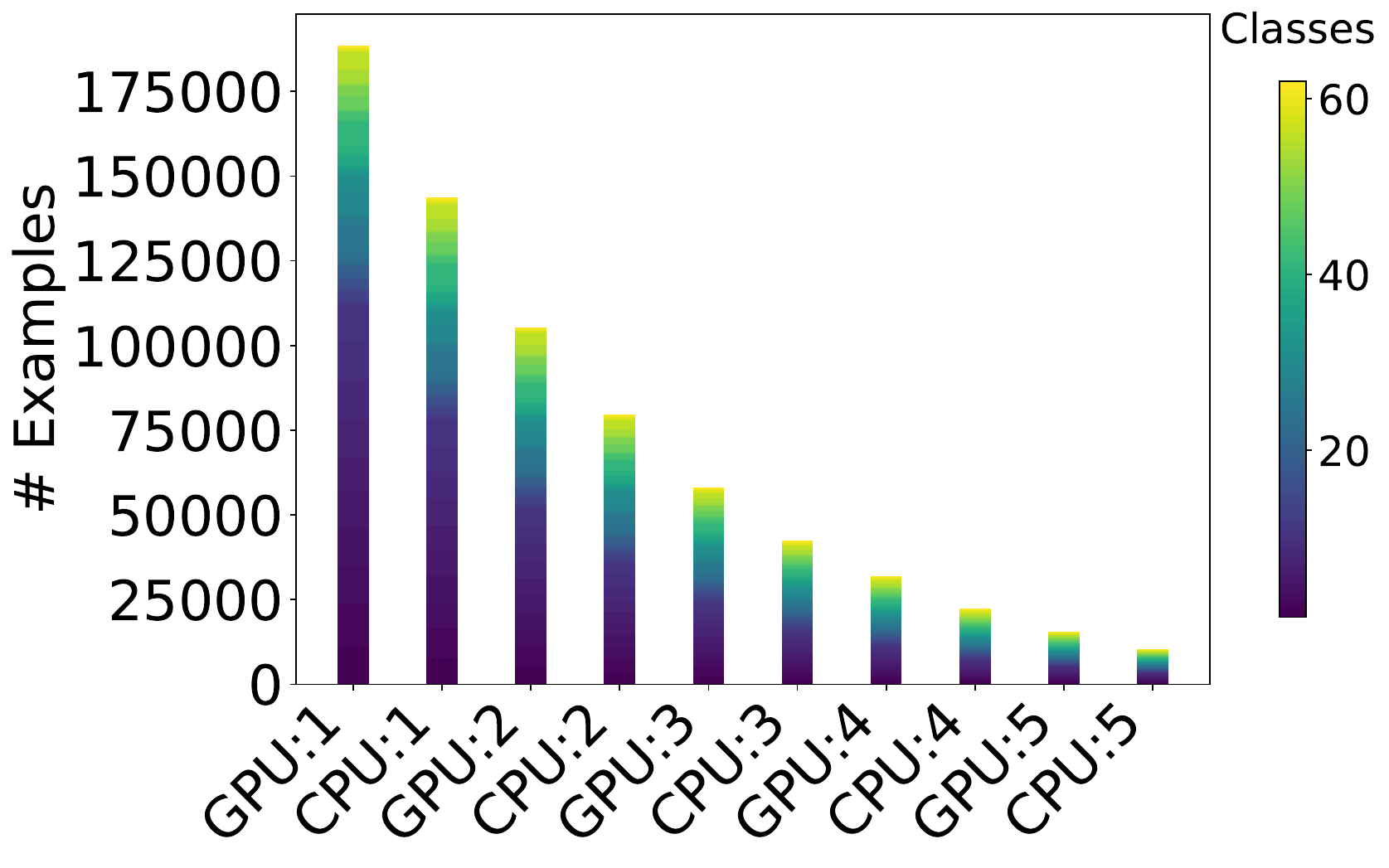}
    \label{subfig:ExtendedMNIST_PowerLaw_IID_DataDistribution}
  }
  \captionsetup{justification=centering}
  \caption{CIFAR and ExtendedMNIST Sample Target Class and Data Size Distributions}
  \label{fig:Cifar10Cifar100ExtendedMNISTDataDistributions}
\end{figure*}

\begin{figure*}[htpb]
  \centering
  \subfloat[Parallel Processing Time]{
    \centering\includegraphics[width=\linewidth]{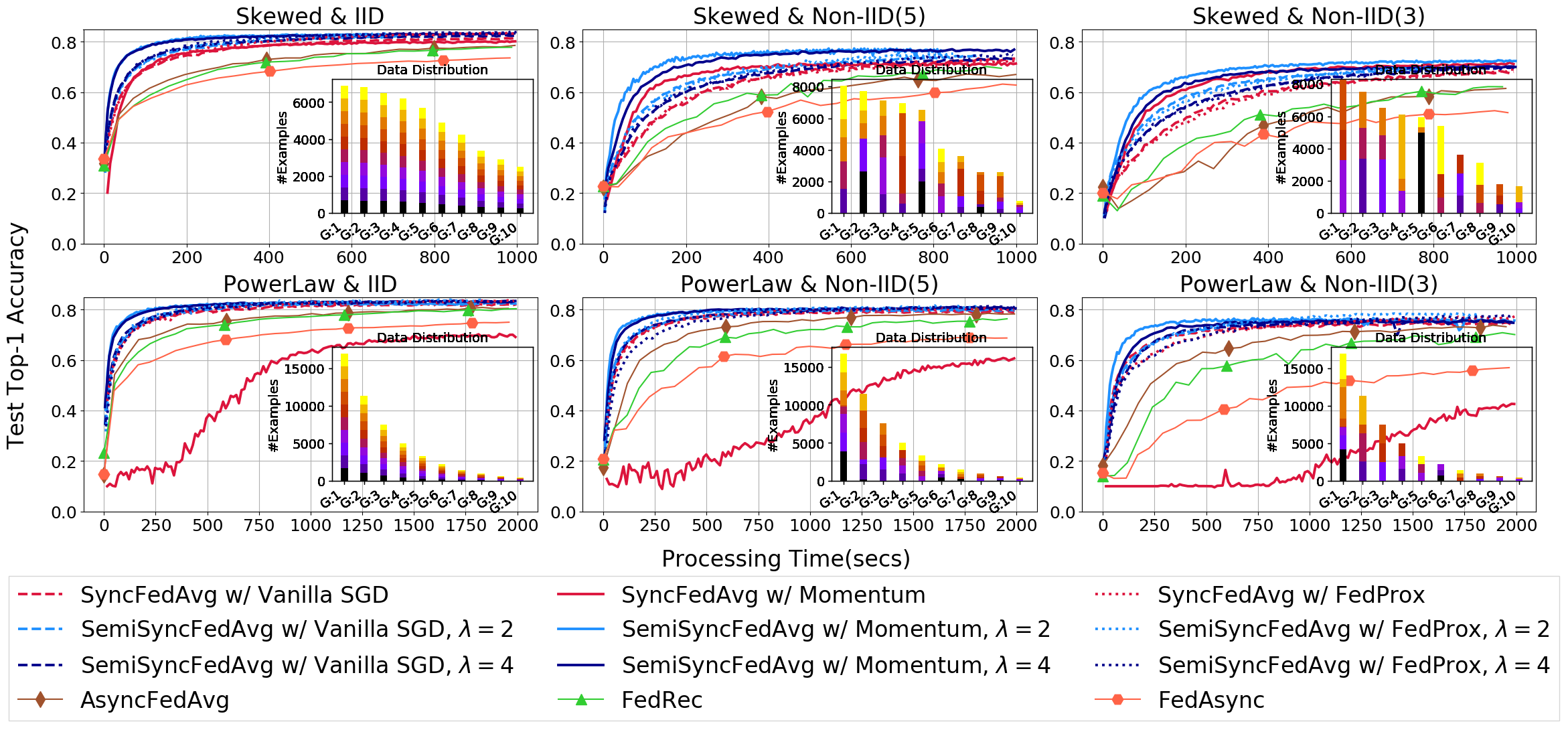}
    \label{subfig:Cifar10_HomogeneousCluster_ProcessingTime}
  }

  \subfloat[Communication Cost (Cumulative Update Requests)]{
    \centering\includegraphics[width=\linewidth]{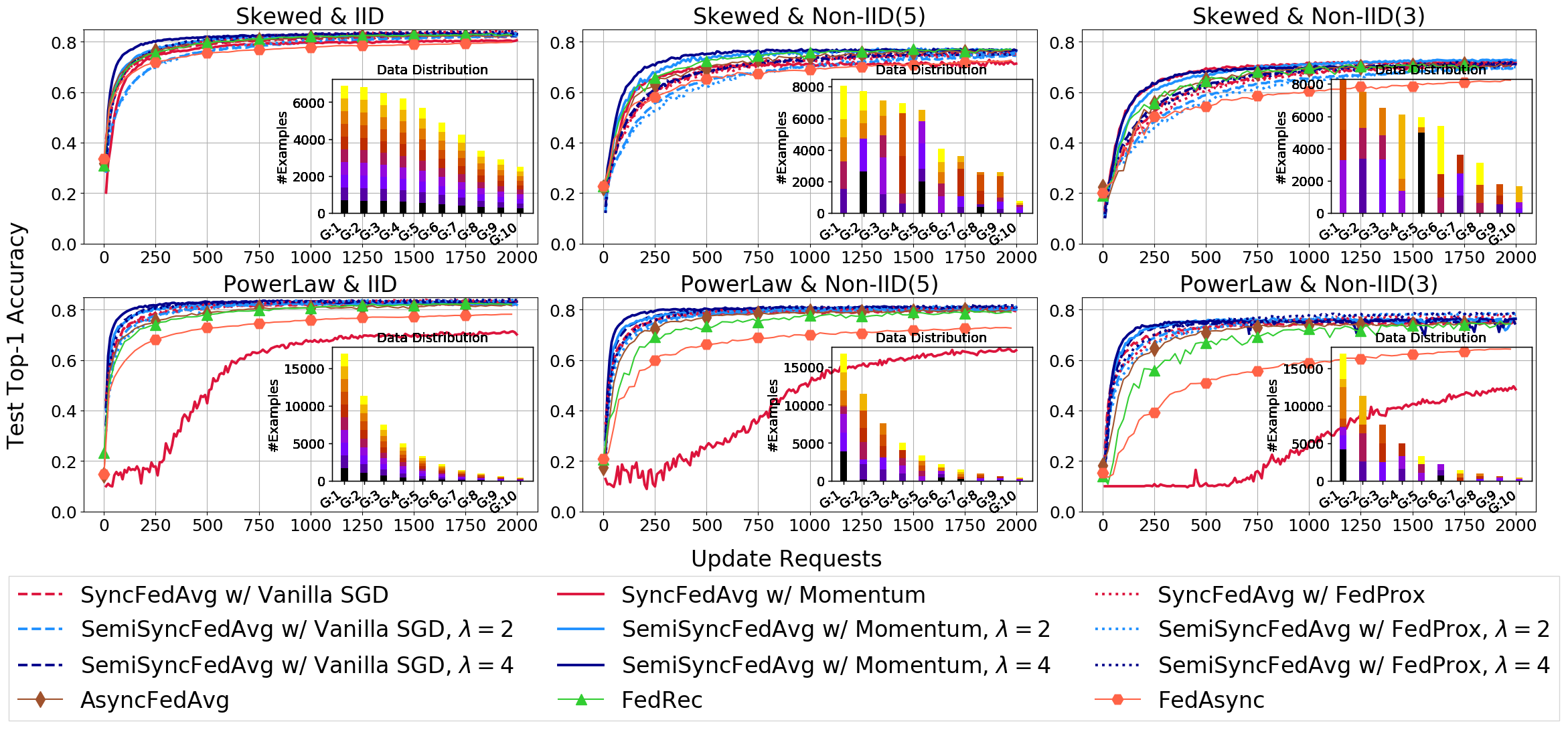}
    \label{subfig:Cifar10_HomogeneousCluster_CommunicationCost}
  }  
  \caption{\textbf{Homogeneous Computational Environment on CIFAR-10}. SemiSync with Momentum ($\lambda=2$) has the fastest convergence. (\textquotesingle G\textquotesingle=GPU in data distribution insets)
  }

  \label{fig:Cifar10HomogeneousCluster}
\end{figure*}

\begin{figure*}[htpb]
  \centering
  \subfloat[Parallel Processing Time]{
    \centering\includegraphics[width=\linewidth]{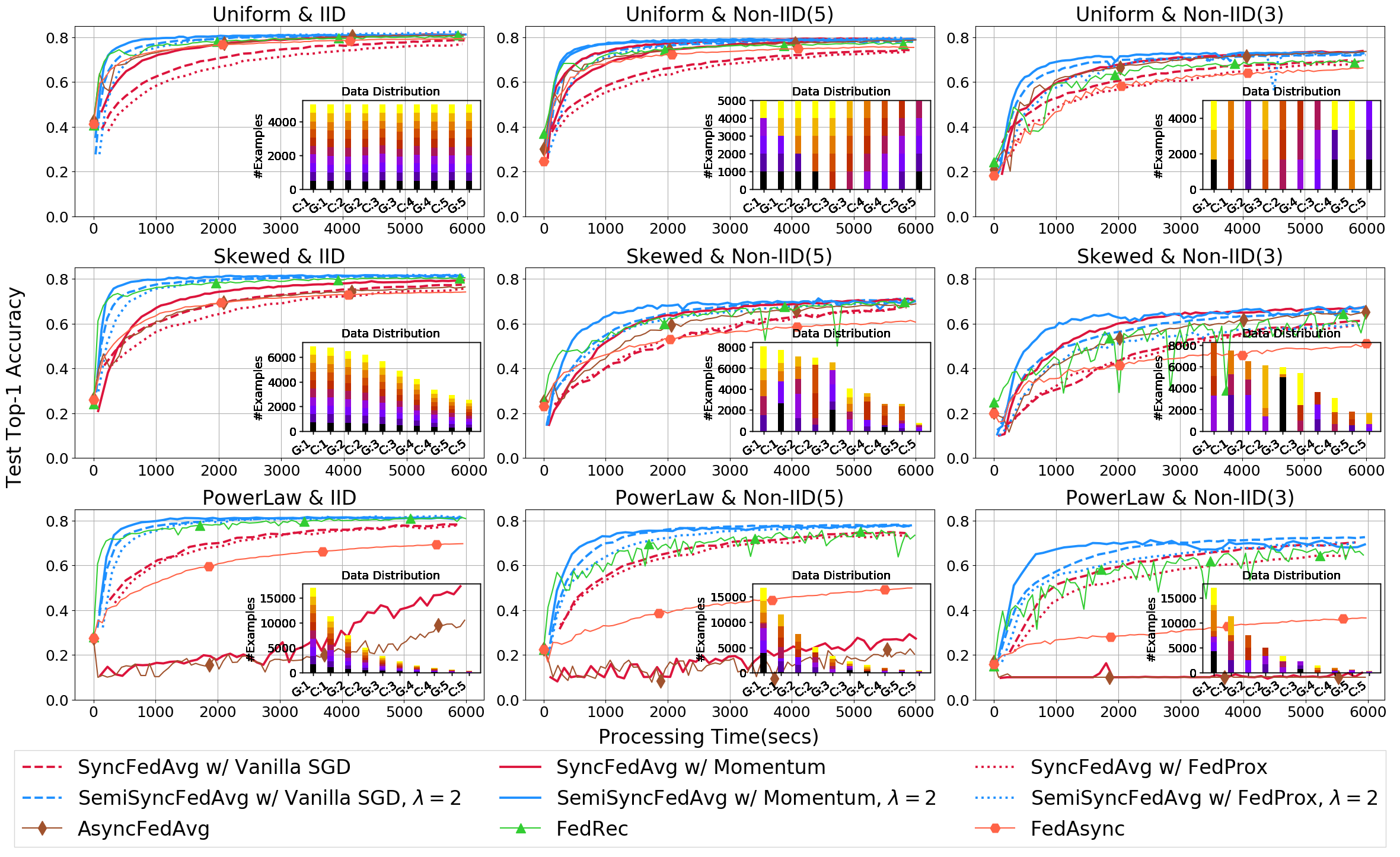}
    \label{subfig:Cifar10_HeterogeneousCluster_ProcessingTime}
  }
  
  \subfloat[Communication Cost (Cumulative Update Requests)]{
    \centering\includegraphics[width=\linewidth]{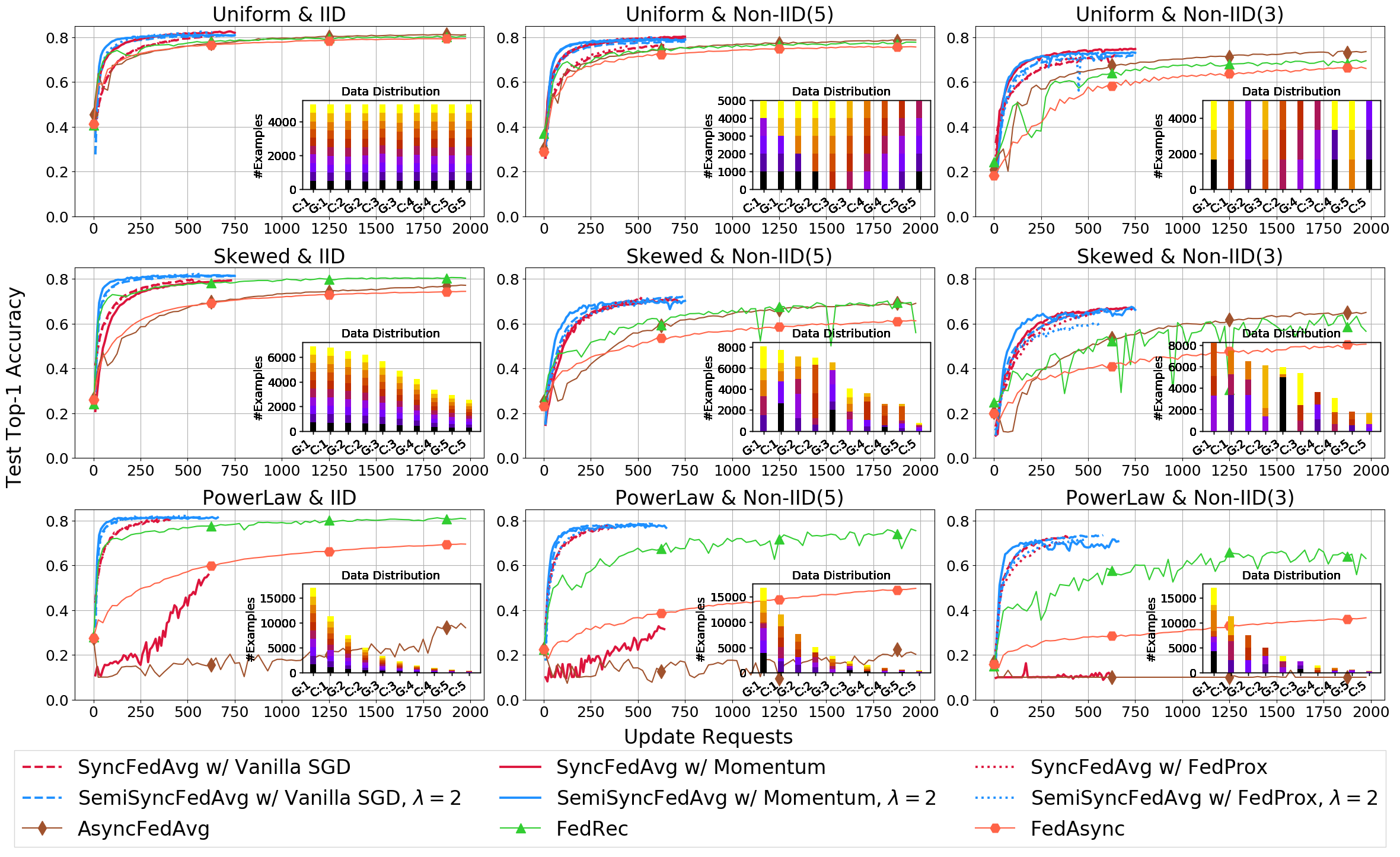}
    \label{subfig:Cifar10_HeterogeneousCluster_CommunicationCost}
  }
    \caption{\textbf{Heterogeneous Computational Environment on CIFAR-10}. SemiSync with Momentum ($\lambda=2$) has the fastest convergence and lowest communication cost for a given accuracy. (\textquotesingle G\textquotesingle=GPU, \textquotesingle C\textquotesingle=CPU)
  }
  \label{fig:Cifar10HeterogeneousCluster}
\end{figure*}

\begin{figure*}[htpb]
  \subfloat[Parallel Processing Time]{
    \includegraphics[width=\linewidth]{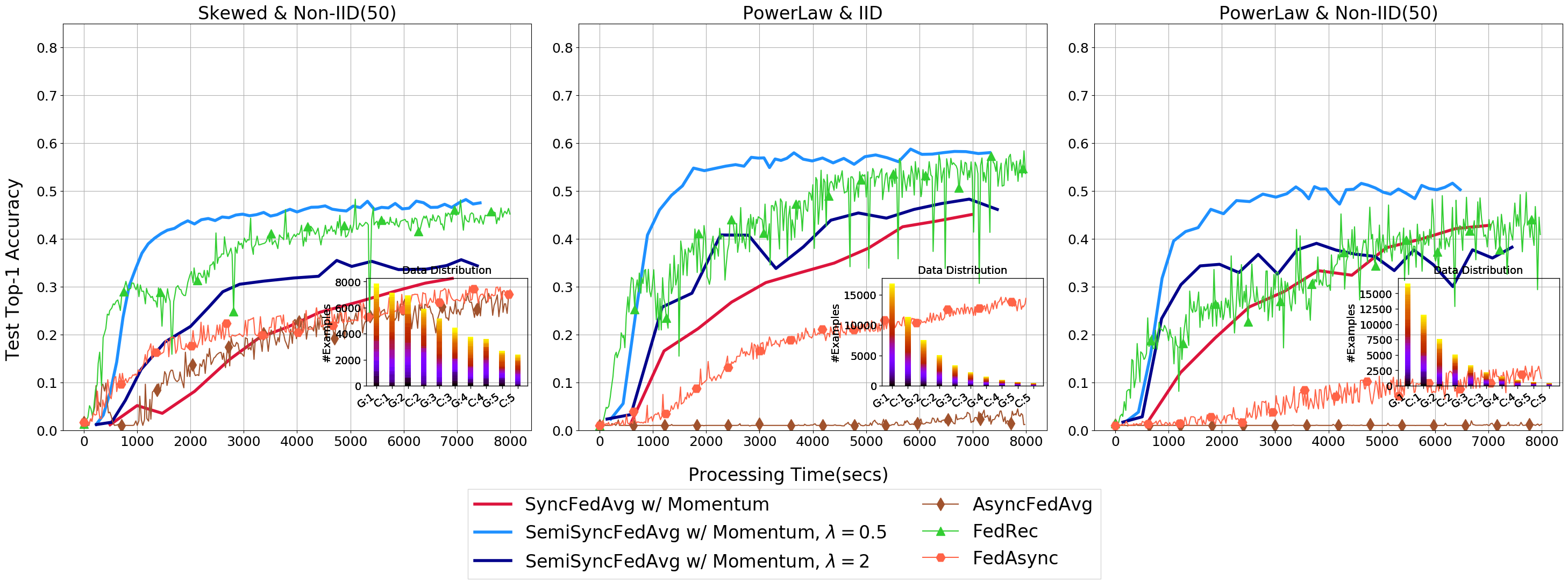}
    \label{subfig:Cifar100_HeterogeneousCluster_ProcessingTime}
  }
  
  \subfloat[Communication Cost (Cumulative Update Requests)]{
    \includegraphics[width=\linewidth]{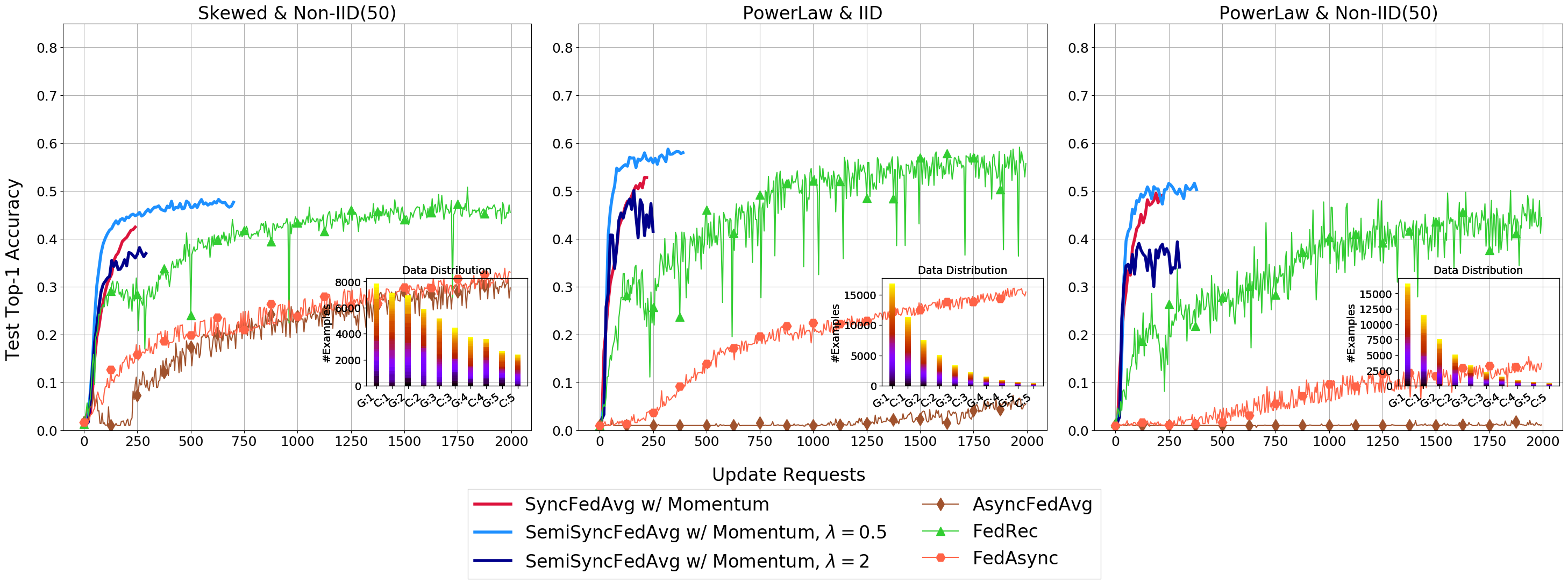}
    \label{subfig:Cifar100_HeterogeneousCluster_CommunicationCost}
  }
    \caption{\textbf{Heterogeneous Computational Environment on CIFAR-100}. SemiSync with Momentum ($\lambda=0.5$) significantly outperforms all other policies in this challenging domain. ('G'=GPU, 'C'=CPU)}
  \label{fig:Cifar100HeterogeneousCluster}
\end{figure*}

\begin{figure*}[htpb]
  \subfloat[Parallel Processing Time]{
    \includegraphics[width=\linewidth]{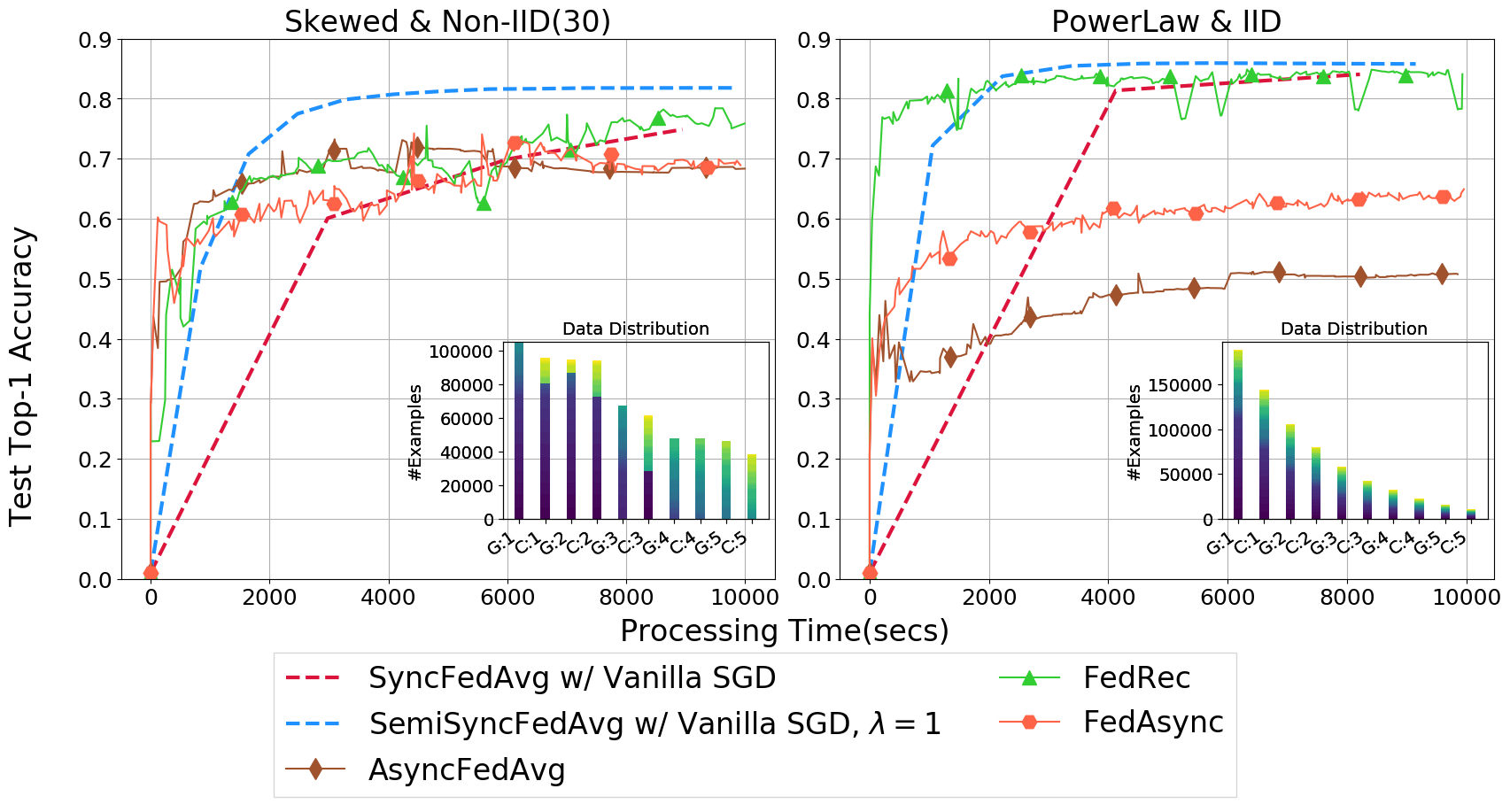}
    \label{subfig:EMNIST_HeterogeneousCluster_ProcessingTime}
  }
  
  \subfloat[Communication Cost (Cumulative Update Requests)]{
    \includegraphics[width=\linewidth]{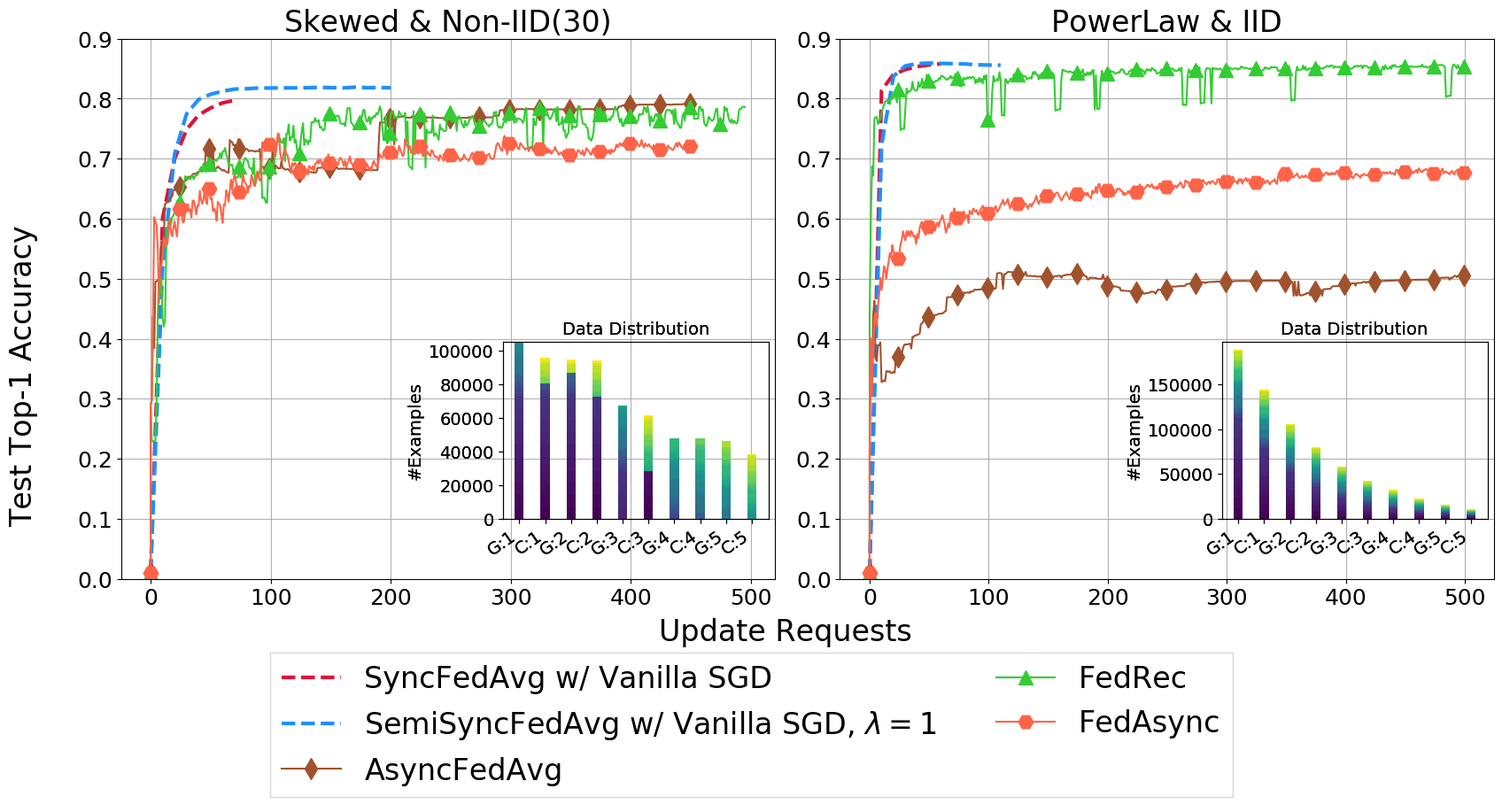}
    \label{subfig:EMNIST_HeterogeneousCluster_CommunicationCost}
  }
    \caption{\textbf{Heterogeneous Computational Environments on ExtendedMNIST By Class}. SemiSync with Vanilla SGD ($\lambda=1$) outperforms all other policies, with faster convergence and less communication cost.}
  \label{fig:EMNISTHeterogeneousCluster}
\end{figure*}

\begin{figure*}[htpb]
  
  \subfloat[Parallel Processing Time]{
    \includegraphics[width=\linewidth]{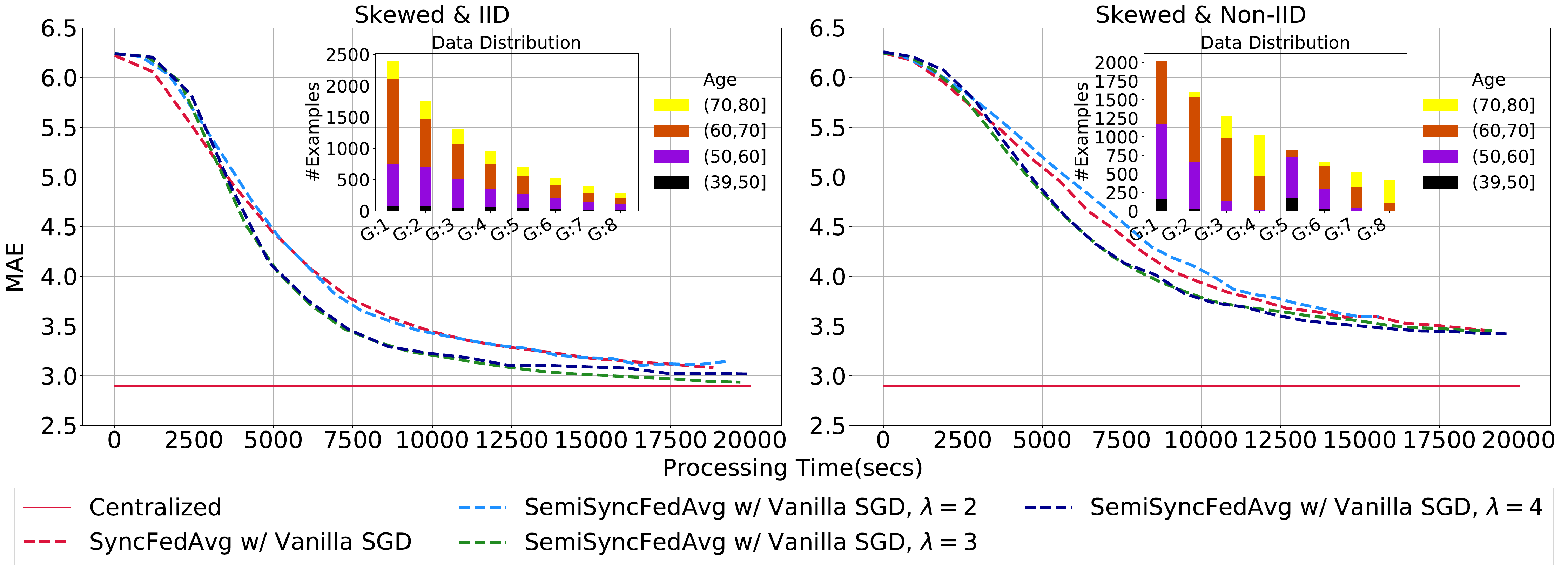}
    \label{subfig:BrainAgeHomogeneousCluster_ProcessingTime}
  }
  
  \subfloat[Communication Cost (Federation Rounds)]{
    \includegraphics[width=\linewidth]{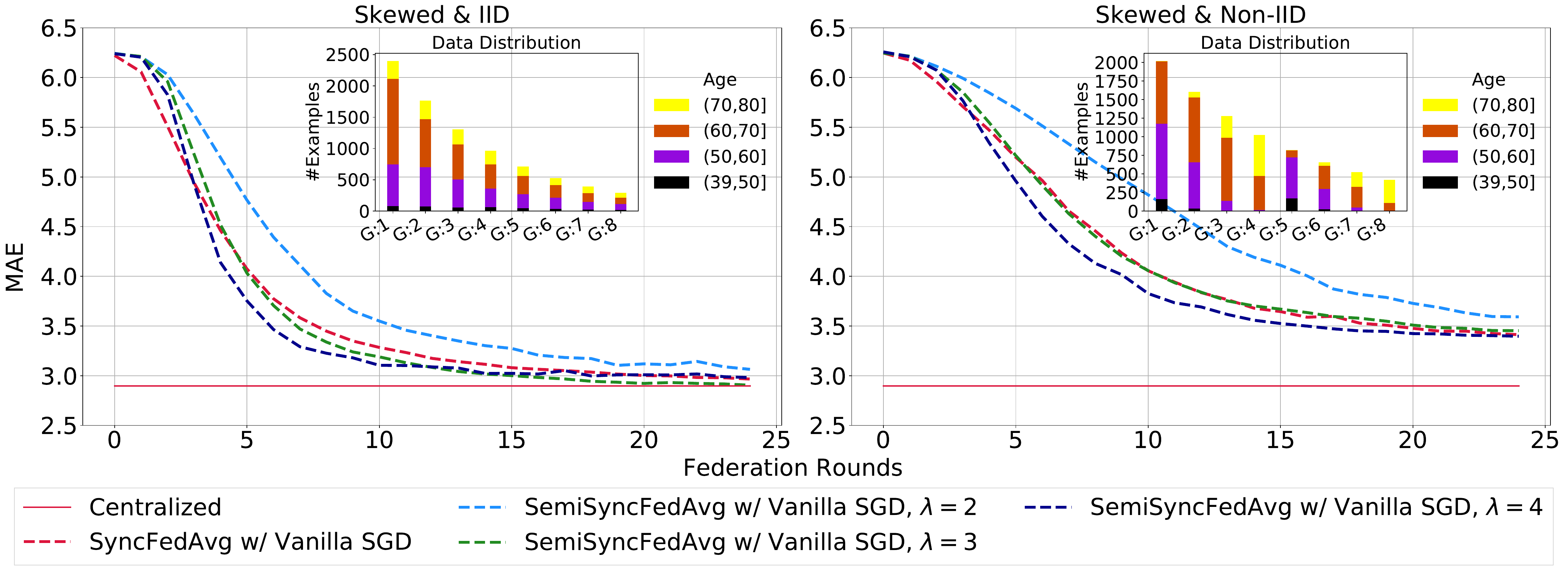}
    \label{subfig:BrainAgeHomogeneousCluster_CommunicationCost}
  }
  \caption{\textbf{Homogeneous Computation, Heterogeneous Data Environments on BrainAge}. SemiSync with Vanilla SGD $\lambda=4$ converges faster and with less communication cost compared to its synchronous counterpart. SemiSync with $\lambda=3$ is close to the performance of the centralized model in Skewed \& IID. 
  }
  \label{fig:BrainAgeHomogeneousCluster}
\end{figure*}

To model \textit{computational heterogeneity}, we use fast (GPU) and slow (CPU) learners. In order to simulate realistic learning environments, we sort each configuration in descending data size order and assign the data to each learner in an alternating fashion (i.e., fast learner, slow learner, fast learner, etc.), except for the uniform distributions where the data size is identical for all learners. Due to space limitations, for every experiment we include the respective data distribution configuration as an inset in the convergence rate plots (Figures~\ref{fig:Cifar10HomogeneousCluster}, \ref{fig:Cifar10HeterogeneousCluster}, \ref{fig:Cifar100HeterogeneousCluster}, \ref{fig:EMNISTHeterogeneousCluster}, \ref{fig:BrainAgeHomogeneousCluster}). Figure \ref{fig:Cifar10Cifar100ExtendedMNISTDataDistributions} shows three representative data distributions for the three classification domains. 

\begin{figure}[tpb]
\raggedright
\begin{minipage}{.3\textwidth}
    \subfloat[Uniform \& IID]{
    \includegraphics[width=\textwidth]{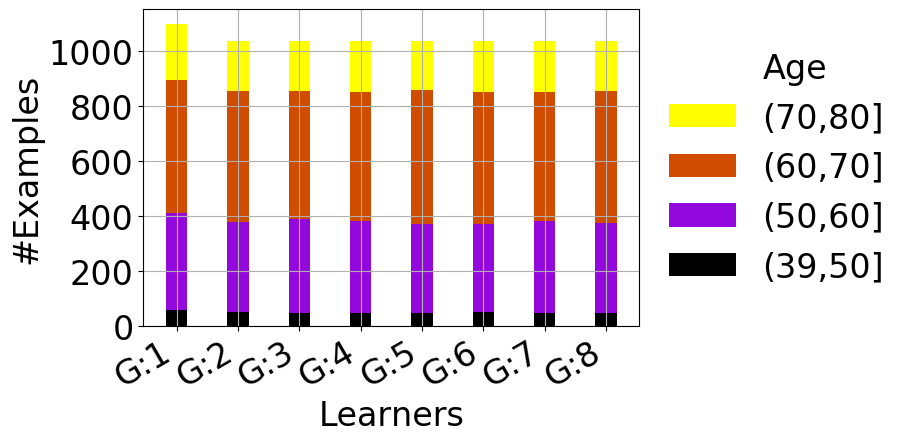}
    \label{subfig:UKBB_Uniform_IID_AgeBucket}
  }
  
  \subfloat[Uniform \& IID]{
    \includegraphics[width=\textwidth]{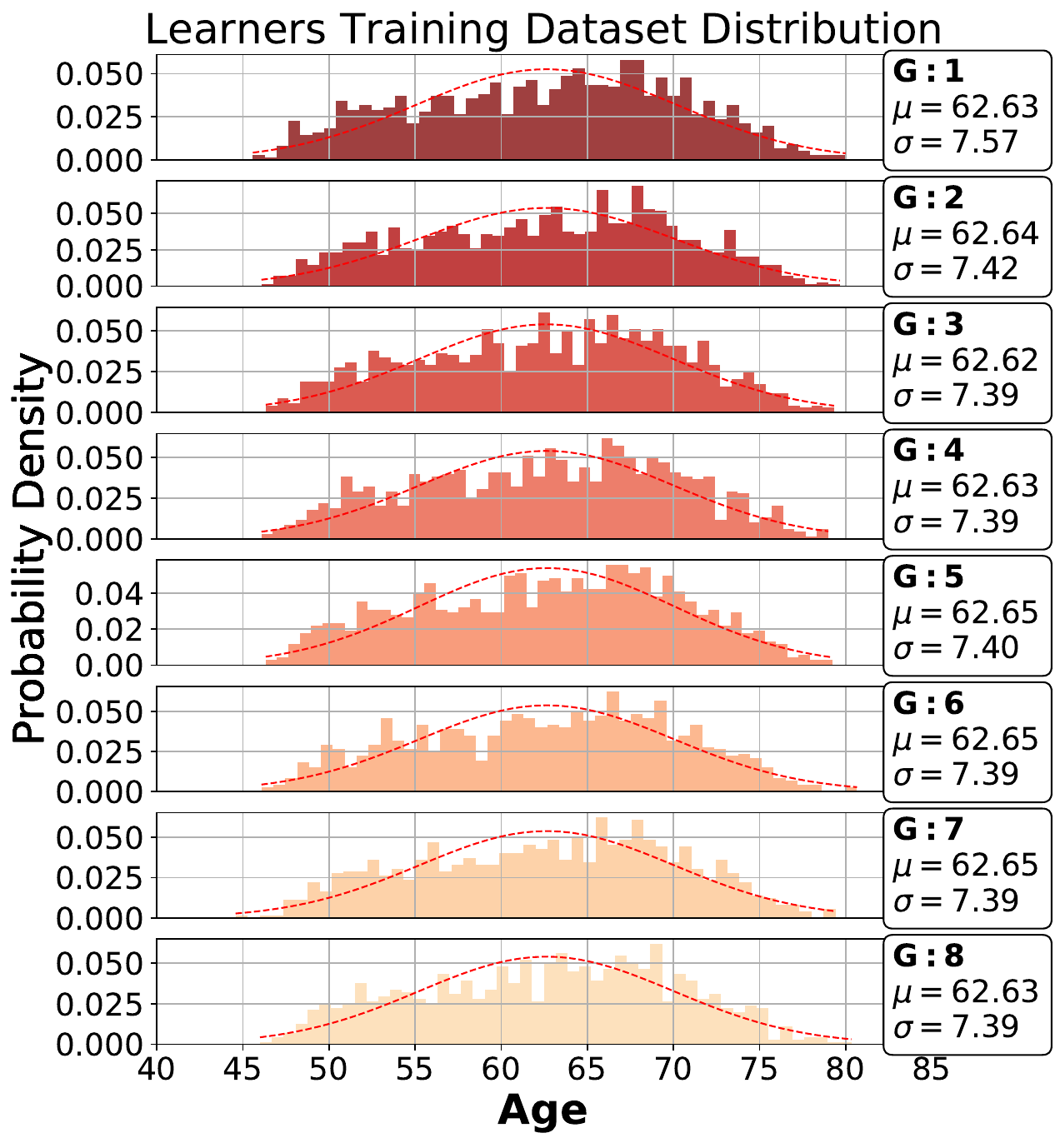}
    \label{subfig:UKBB_Uniform_IID_AgeDistribution}
  }
  \label{fig:UKBB_Uniform_IID}
\end{minipage}%
\begin{minipage}{.3\textwidth}

  \subfloat[Skewed \& IID]{
    \includegraphics[width=\textwidth]{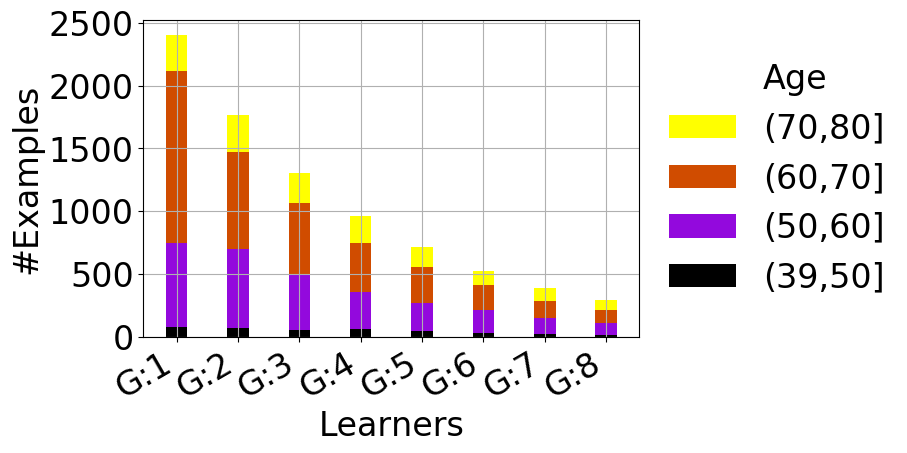}
    \label{subfig:UKBB_Skewed_IID_AgeBucket}
  }
  
  \subfloat[Skewed \& IID]{
    \includegraphics[width=\textwidth]{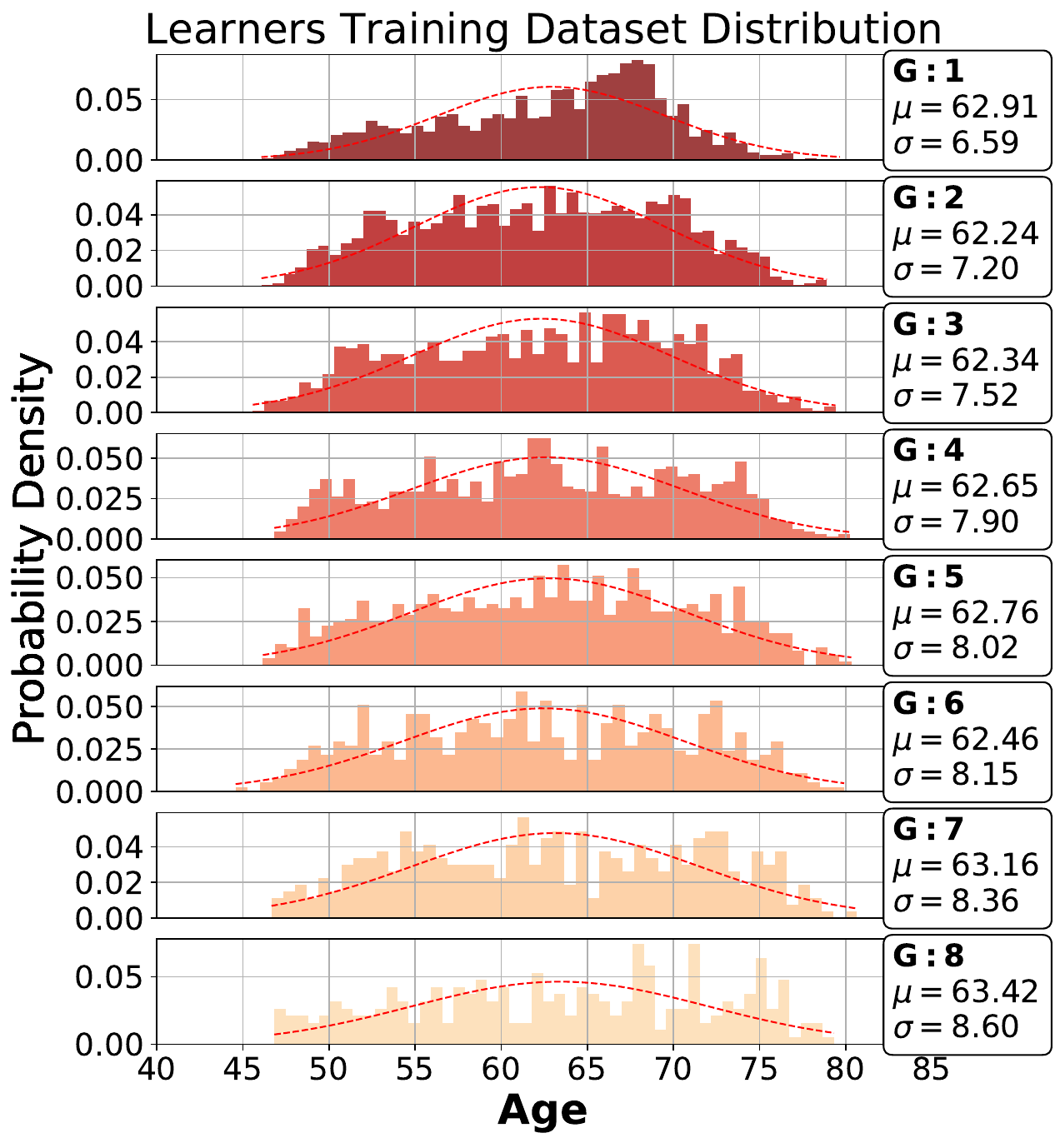}
    \label{subfig:UKBB_Skewed_IID_AgeDistribution}
  }
  \label{fig:UKBB_Skewed_IID}
\end{minipage}
\begin{minipage}{.3\textwidth}
  \subfloat[Skewed \& Non-IID]{
    \includegraphics[width=\textwidth]{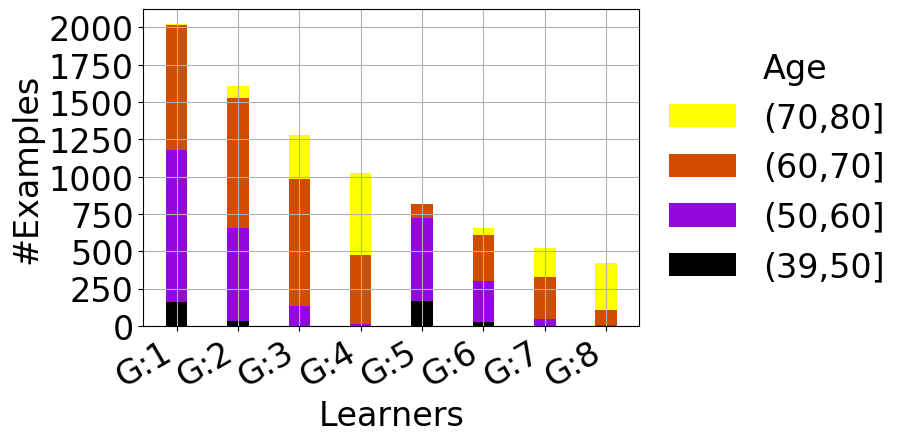}
    \label{subfig:UKBB_Skewed_NonIID_AgeBucket}
  }
  
  \subfloat[Skewed \& Non-IID]{
    \includegraphics[width=\textwidth]{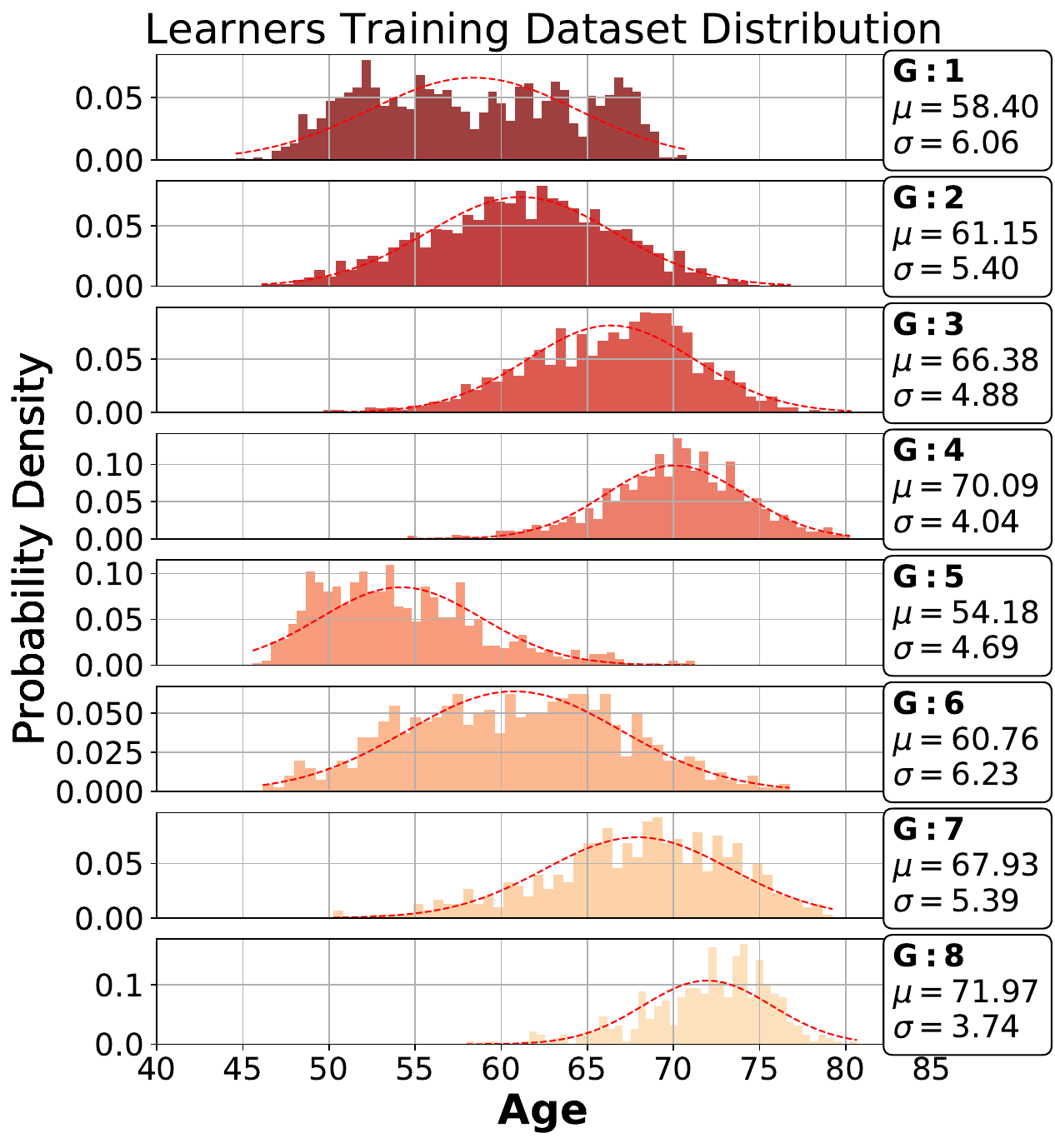}
    \label{subfig:UKBB_Skewed_NonIID_AgeDistribution}
  }
  \label{fig:UKBB_Skewed_NonIID}
\end{minipage}
\caption{UK Biobank Data Distributions. \textbf{Top Row:} Amount of data examples across learners in terms of age buckets/ranges. \textbf{Bottom Row:} Local age distribution (histogram) of each learner.}
\label{fig:UKBB_DataDistributions}
\end{figure}

The data distributions of the classification tasks that we investigate are based on the work of~\cite{zhao2018federated}, where the data are evenly (i.e, Uniform in our case) partitioned across 10 learners, and with different class distribution per learner (i.e., Non-IID(2) refers to examples from 2 classes per learner). We extend their work by also investigating the cases where the size of the partitions is not uniform, but follows a skewed or a power law distribution (called quantity skew/unbalanceness in~\cite{kairouz2019advances}).

For the statistical heterogeneity of the regression task in BrainAge, we partition (following~\cite{stripelis2021scaling}) the UKBB neuroimaging dataset~\cite{miller2016multimodal} into training (8356 records) and test (2090 records) across a federation of 8 learners with skewed data amounts, and IID and Non-IID age distributions. In the IID case, every learner holds data examples from all possible age ranges, while in the Non-IID case learners hold examples from a subset of age ranges. Figure \ref{fig:UKBB_DataDistributions} shows the distribution of the training examples across the 8 learners over three different learning environments in BrainAge. Every environment is evaluated on the same test dataset representative of the global age distribution.

Within each learning domain, all training policies were run for the same amount of wall-clock time. 
In homogeneous computational environments (cf. Figure \ref{subfig:Cifar10_HomogeneousCluster_CommunicationCost}), the total number of update requests for synchronous and asynchronous policies is similar, since any latency occurs only due to differences in the amount of training data. However, in heterogeneous computational environments, during asynchronous training, computationally fast learners issue many more update requests than slow learners. In synchronous and semi-synchronous policies, the update requests are substantially smaller and driven by the slowest learner in the federation. To highlight the differences in communication costs, Figures \ref{subfig:Cifar10_HeterogeneousCluster_CommunicationCost}, \ref{subfig:Cifar100_HeterogeneousCluster_CommunicationCost} and \ref{subfig:EMNIST_HeterogeneousCluster_CommunicationCost} show the number of update requests issued by all learners \textit{for the same wall-clock time period} (set in Figures
\ref{subfig:Cifar10_HeterogeneousCluster_ProcessingTime}, \ref{subfig:Cifar100_HeterogeneousCluster_ProcessingTime} and \ref{subfig:EMNIST_HeterogeneousCluster_ProcessingTime}, respectively).
Point markers in the asynchronous policies are just a visual aid to distinguish them from synchronous and semi-synchronous policies.

\paragraph{CIFAR-10 Results}
Figure~\ref{fig:Cifar10HomogeneousCluster} shows the performance of synchronous, semi-synchronous, and asynchronous policies in a \textit{homogeneous} computational environment, where all learners have the same computational capabilities (10 GPUs). We compare the different policies on heterogeneous amounts of data per learner (Skewed, Power Law) since for a homogeneous computational environment with an equal number of data points (Uniform) across learners, all policies are equivalent. For the synchronous policies, idle time still occurs due to the different data amount in the Skewed and Power Law data distributions. For SemiSync and asynchronous policies, there is no idle time. 
We evaluate SyncFedAvg and SemiSyncFedAvg with Vanilla SGD, Momemtum SGD, and FedProx, and AsyncFedAvg, FedRec, and FedAsync with MomentumSGD.
SyncFedAvg with Momentum converges very slowly in Power Law data distributions, but using FedProx as a local optimizer rescues it. 
SemiSync with Momentum and $\lambda=2$ results in the fastest convergence across the experiments in Figure \ref{fig:Cifar10HomogeneousCluster}(a) (see also Table \ref{tbl:Cifar10HomogeneousCluster}). 
Similar results hold for communication cost in terms of update requests as shown in Figure~\ref{fig:Cifar10HomogeneousCluster}(b). 
When the amounts of data across learners is not too great (Skewed data distributions) AsyncFedAvg and FedRec have comparable performance, but as the difference in data increases (PowerLaw data distributions), AsyncFedAvg is more efficient compared to the staleness-aware weighting scheme of both FedRec and FedAsync.

Figure \ref{fig:Cifar10HeterogeneousCluster} shows the performance on the CIFAR-10 domain of synchronous, asynchronous, and semi-synchronous policies in a \textit{heterogeneous} computational environment (with 10 learners: 5 fast GPUs, and 5 slow CPUs; the CPUs' batch processing is 10 times slower than the GPUs). 
Again, our SemiSync with Momentum ($\lambda=2$) has the best performance with faster convergence  (some other policies reach comparable accuracy levels eventually). As we move towards more challenging Non-IID learning environments, all methods show a reduction in the final performance (lower accuracy). 
SyncFedAvg with Momentum performs reasonably well with moderate levels of data heterogeneity (Skewed data amounts, with either IID or Non-IID distributions). However, in more extreme data distributions (Power Law), it learns very slowly, since the local models are more disparate and the momentum factor limits changes to the community model. In SemiSync the fast learners perform more local iterations and mix more frequently, which facilitates the convergence of the federation even for the same momentum factor that slowed learning for SyncFedAvg. This is exarcerbated in the heterogeneous case, compared to the homogeneous case, since the training speed differences are much greater.
Interestingly, SyncFedAvg with FedProx performs much better in these cases, although still worse than the SemiSync policies. 
FedRec dominates other asynchronous policies on convergence rate, especially in IID environments and in Power Law distributions. In Uniform and Skewed Non-IID environments, its performance is comparable to AsyncFedAvg.
Comparing the experiments based on data amounts, we can see that in some learning environments, such as Skewed \& Non-IID(5) and Power Law \& Non-IID(5) in Figures \ref{fig:Cifar10HomogeneousCluster} and \ref{fig:Cifar10HeterogeneousCluster}, the optimal accuracy is higher in the Power Law case compared to its Skewed counterpart. 
This is due to the fact that the head (G:1) of the Power Law distribution covers most of the domain data (8 classes) and therefore its local model is more valuable and has a greater contribution value in the community model.

The communication cost of SemiSync policies is comparable to synchronous policies reaching a high accuracy very quickly, while asynchronous policies require many more update requests to achieve the same (or worse) level of accuracy (Figure~\ref{fig:Cifar10HeterogeneousCluster}(b)). For Power Law data distributions SyncFedAvg with Momentum fails to learn, but SyncFedAvg with FedProx learns and efficiently uses communication. 
Similar to the homogeneous case, our semi-synchronous policies dominate in heterogeneous environments (see also Table \ref{tbl:Cifar10HeterogeneousCluster}).

\paragraph{CIFAR-100 Results}
Figure~\ref{fig:Cifar100HeterogeneousCluster} shows the performance on the CIFAR-100 domain of synchronous, asynchronous, and semi-synchronous policies in a heterogeneous computational environment (with 10 learners: 5 fast GPUs, and 5 slow CPUs). Due to the size and computational complexity of the ResNet-50 model used to train on the CIFAR-100 domain, the performance difference between fast and slow learners is large (CPUs batch processing is 33 times slower than the GPUs). Using a smaller value of $\lambda$ leads to better results for the SemiSync policy. We use $\lambda=0.5$, which means that the slowest learner processes only half of its local dataset at each SemiSync synchronization point. However, since the batches are chosen randomly, after two synchronization points all the data is processed (on average).
FedRec dominates alternative asynchronous policies, though is less stable. 
Our SemiSync with Momentum ($\lambda=0.5$) policy yields the fastest convergence rate and final accuracy, while remaining communication efficient.

\paragraph{ExtendedMNIST Results}
Figure~\ref{fig:EMNISTHeterogeneousCluster} shows the results on the ExtendedMNIST By Class domain on heterogeneous environments (10 learners: 5 fast, 5 slow; slow learners (CPUs) are 16 times slower than fast learners (GPUs)). 
Extended MNIST By Class is a very challenging learning task due to its unbalanced distribution of target classes and large amount of data samples (731,668 training examples). As we empirically show, SemiSync performs considerably better compared to other policies both in terms of convergence speed and in terms of communication cost. Among the asynchronous policies, FedRec has faster convergence and better generalization. 
Interestingly, asynchronous policies have faster convergence at the beginning of the federated training, a behavior that is more pronounced for FedRec in the Power Law \& IID environment. Given the large number of records allocated to the slow learners (e.g., C:1 owns $\sim$90,000 in Skewed and $>$140,000 examples in Power Law), the total processing time required to complete their local training task is much higher compared to fast. Moreover, since in asynchronous policies no idle time exists for the fast learners, the convergence of the community model is driven by their learning pace and more frequent communication, whereas in synchronous and semi-synchronous all learners need to complete their local training task before a new community model is computed; hence the delayed convergence of Sync and SemiSync at the start of training.

\paragraph{BrainAge Results}
We evaluate the synchronous and SemiSync policies on a computationally homogeneous environment with Skewed IID and Non-IID data distributions (for Uniform data amounts both policies behave identically, see ~\cite{stripelis2021scaling} for results in this environment) on the BrainAge domain.
In terms of parallel processing time, SemiSync with $\lambda=3,4$ provides faster convergence for both Skewed IID and Non-IID distributions compared to the synchronous policy. In terms of communication cost, SemiSync with $\lambda=4$ is more communication efficient for both data distributions, with $\lambda=3$ being comparable to synchronous. For IID, SemiSync with $\lambda=3$ leads to the smallest mean absolute error, which is very close to the error reached by the centralized model.

\begin{figure}[htbp]
  \subfloat[Uniform \& IID] {
    \includegraphics[width=0.45\linewidth]{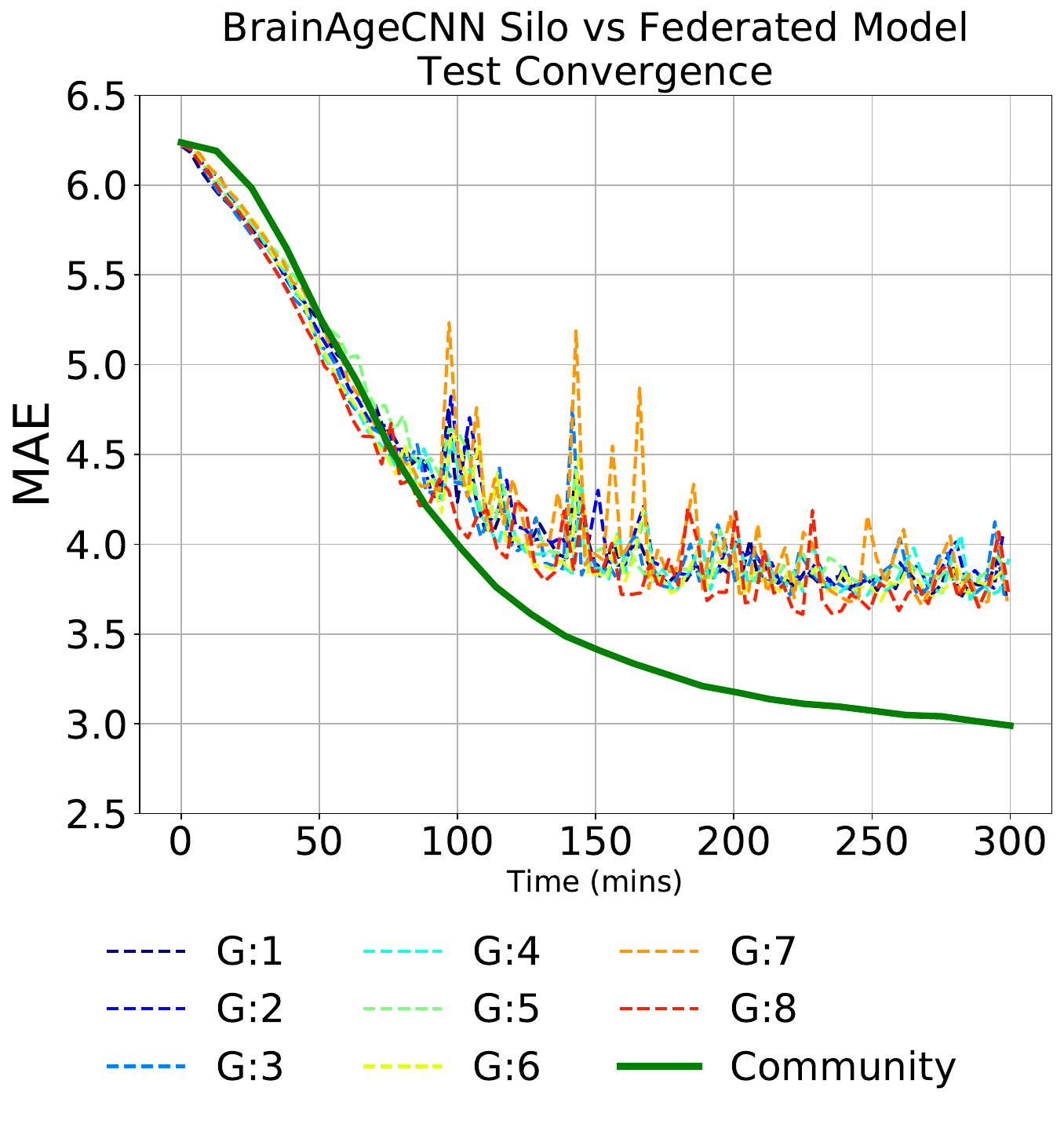}
    \label{subfig:Silos_vs_Federation_UniformIID}
  }
  \subfloat[Skewed \& Non-IID]{
    \includegraphics[width=0.45\linewidth]{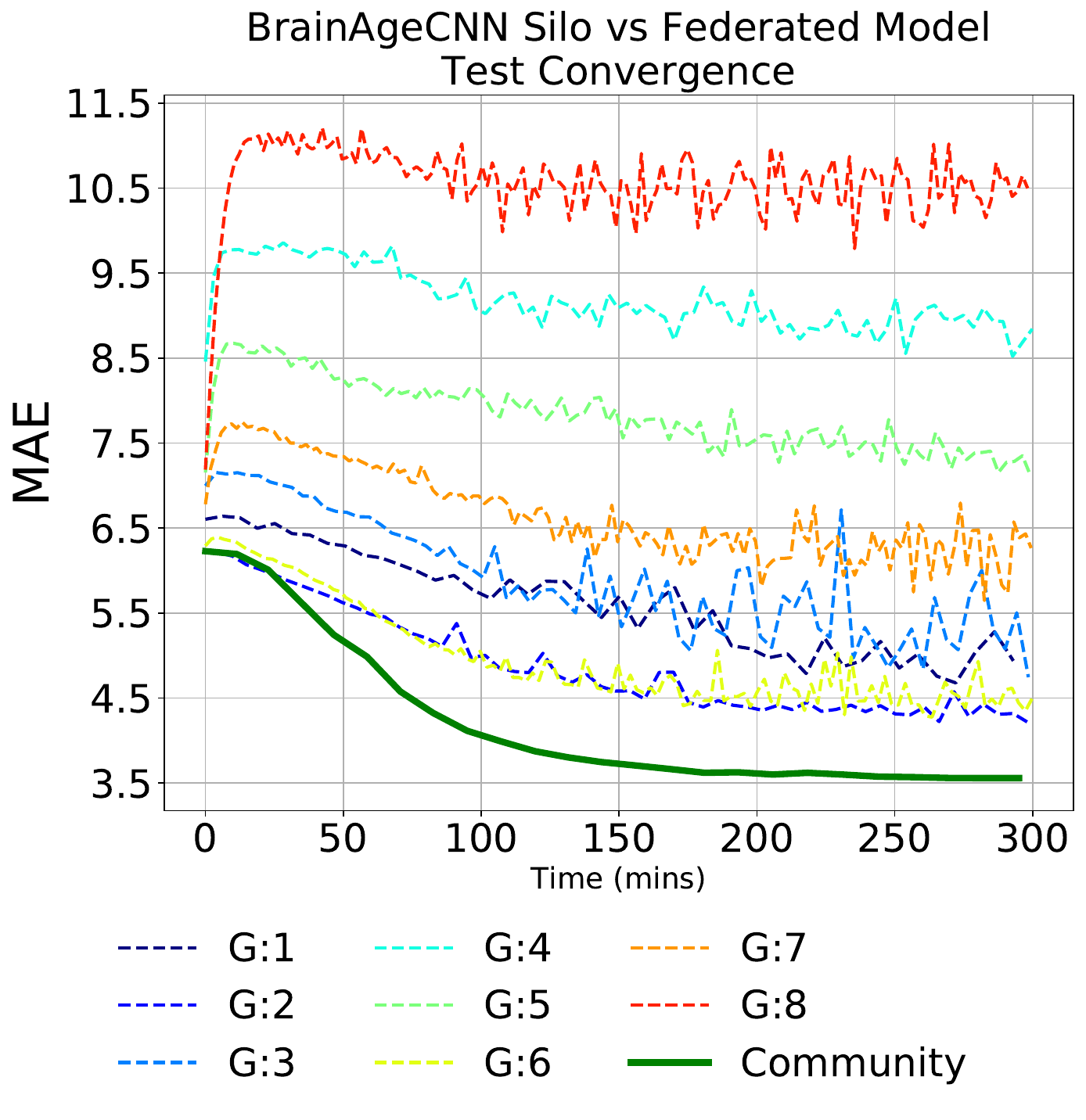}
    \label{subfig:Silos_vs_Federation_SkewedNonIID}
  }
  \caption{Performance evaluation of standalone silo against the federation (community) model on the BrainAge learning task. Federated learning leverages all available data points and learns a more generalized model.}
  \label{fig:Silos_vs_Federation}
\end{figure}

\paragraph{Models Performance: Time, Communication, and Energy}
We analyze the performance of the different federated training policies in terms of time, communication,%
\footnote{We disregard model transmission costs. In  heterogeneous environments for CIFAR-10, CIFAR-100 and ExtendedMNIST, the model size, average model transmission time, average processing time for SemiSync, and ratio of model transmission to computation are: (4MB, 0.4secs, 80secs $\lambda=2$, 0.005),
(5MB, 0.4secs, 80secs $\lambda=2$, 0.003), and (50MB, 5secs, 1000secs $\lambda=1$, 0.005), respectively. Similarly, for the homogeneous environment of BrainAge we have (11MB, 0.9secs, 850secs $\lambda=2$, 0.001).}
 and energy costs, in the CIFAR-10 domain in both homogeneous (Table~\ref{tbl:Cifar10HomogeneousCluster}) and heterogeneous (Table~\ref{tbl:Cifar10HeterogeneousCluster}) environments.
To compare the training policies, we pick a target accuracy that can be reached by (most of) the policies and calculate the different metrics as defined in Section~\ref{subsec:TrainingPoliciesCostAnalysis}.
To compute the energy cost in the heterogeneous computational environment, we set $\epsilon_k = 2$ for GPUs and $\epsilon_k = 1$ for CPUs (see eq. \ref{eq:EnergyCostFunction}). Since estimating the full energy consumption (network, storage, power conversion) is challenging, we compute the energy cost ratio between GPUs and CPUs based on their Thermal Design Power (TDP) value~\cite{dayarathna2015data}, which accounts for the maximum amount of heat generated by a processor. For the GPU GeForce GTX 1080 Ti the TDP value is $\sim$180W, while for the Intel(R) Xeon(R) CPU the TDP value is $\sim$90W.
In both homogeneous and heterogeneous domains, SemiSync with Momentum ($\lambda=2$) performs best in terms of the (parallel/wall-clock) time to reach the desired target accuracy and in terms of the energy cost (with depends on the cumulative processing time across all learners needed to reach that accuracy). In heterogeneous domains, SemiSync with Momentum ($\lambda=2$) is also the most efficient in terms of communication. In homogeneous domains SemiSync with Momentum ($\lambda=4$) is more communication efficient, but with slightly slower convergence and larger energy cost, when compared to ($\lambda=2$). Remarkably, when our SemiSync strategy is combined with any local optimizer and compared to its synchronous counterpart, it yields significant energy savings close to an average 40\% energy cost reduction in both homogeneous and heterogeneous computational environments (cf. Tables \ref{tbl:Cifar10HomogeneousCluster} and \ref{tbl:Cifar10HeterogeneousCluster}). 
Finally, comparing the performance of the different local optimizers, Momentum SGD provides accelerated convergence with a lower communication overhead compared to Vanilla SGD and FedProx. At the same time Momentum SGD is the most energy-efficient policy, compared to the Synchronous Vanilla SGD baseline, with a reduction of 3 to 9 times the energy cost.

\begin{table}[htpb]
\noindent
\small
\setlength\tabcolsep{2pt}
    \centering
    \begin{tabular}{@{}clccccc@{}}
    \multicolumn{1}{c}{\textbf{\begin{tabular}[c]{@{}c@{}}Experiment\end{tabular}}} &
    \multicolumn{1}{c}{\textbf{\begin{tabular}[c]{@{}c@{}}Policy\end{tabular}}} &
    \multicolumn{1}{c}{\textbf{\begin{tabular}[c]{@{}c@{}}Iterations\end{tabular}}} &
    \multicolumn{1}{c}{\textbf{\begin{tabular}[c]{@{}c@{}}Parallel\\Time(s)\end{tabular}}} &
    \multicolumn{1}{c}{\textbf{\begin{tabular}[c]{@{}c@{}}Cumulative\\Time(s)\end{tabular}}} &
    \multicolumn{1}{c}{\textbf{\begin{tabular}[c]{@{}c@{}}Com.\\Cost\end{tabular}}} &
    \multicolumn{1}{c}{\textbf{\begin{tabular}[c]{@{}c@{}}Energy Cost\\(Efficiency)\end{tabular}}}
    \\
    \toprule
    \parbox[t]{2mm}{\multirow{10}{*}{\rotatebox[origin=c]{90}{\begin{tabular}[c]{@{}c@{}}Skewed \& Non-IID(5)\\0.7\end{tabular}}}}
    & Sync w/ Vanilla & 123220 & 613 & 6133 & 610 & 12267\\
    & SemiSync ($\lambda=2$) w/ Vanilla & 156025 & 444 & 3541 & 970 & 7083 (1.7x) \\
    & SemiSync ($\lambda=4$) w/ Vanilla & 201385 & 551 & 4507 & 630 & 9014 (1.3x)\\
    \cmidrule{2-7}
    & Sync w/ Momentum & 92920 & 319 & 3190 & 460 & 6381 (1.9x)\\
    & SemiSync ($\lambda=2$) w/ Momentum & 47485 & \textbf{121} & \textbf{1005} & 300 & \textbf{2010 (6.1x)}\\
    & SemiSync ($\lambda=4$) w/ Momentum & 58825 & 161 & 1324 & \textbf{190} & 2648 (4.6x) \\
    \cmidrule{2-7}
    & Sync w/ FedProx & 135340 & 630 & 6305 & 670 & 12610 (0.9x) \\
    & SemiSync ($\lambda=2$) w/ FedProx & 144685 & 380 & 3134 & 900  & 6268 (1.9x)\\
    & SemiSync ($\lambda=4$) w/ FedProx & 185185 & 481 & 3876 & 580 & 7752 (1.5x)\\
    \cmidrule{2-7}
    & FedRec & 61224 & 924 & 3830 & 358 & 7661 (1.6x) \\
    & AsyncFedAvg & 78098 & 1166 & 4864 & 456 & 9729 (1.2x) \\
    & FedAsync & 198990 & 3120 & 12709 & 1164 & 25419 (0.4x) \\
    \midrule
    \parbox[t]{2mm}{\multirow{10}{*}{\rotatebox[origin=c]{90}{\begin{tabular}[c]{@{}c@{}}Power Law \& Non-IID(3)\\0.65\end{tabular}}}}
    & Sync w/ Vanilla & 24240 & 181 & 1811 & 120 & 3623\\
    & SemiSync ($\lambda=2$) w/ Vanilla & 54905 & 199 & 1347 & 170 & 2695 (1.3x)\\
    & SemiSync ($\lambda=4$) w/ Vanilla & 68505 & 208 & 1694 & 110 & 3389 (1.1x)\\
    \cmidrule{2-7}
    & Sync w/ Momentum & \multicolumn{5}{c}{\textit{did not reach target accuracy}} \\
    & SemiSync ($\lambda=2$) w/ Momentum & 31105 & \textbf{86} & \textbf{667} & 100 & \textbf{1335 (2.7x)}\\
    & SemiSync ($\lambda=4$) w/ Momentum & 48105 & 146 & 1181 & \textbf{80} & 2363 (1.5x)\\
    \cmidrule{2-7}
    & Sync w/ FedProx & 38380 & 287 & 2876 & 190 & 5752\\
    & SemiSync ($\lambda=2$) w/ FedProx & 75305 & 208 & 1620 & 230 & 3241 (1.1x)\\
    & SemiSync ($\lambda=4$) w/ FedProx & 95705 & 291 & 2157 & 150 & 4315 (0.8x)\\
    \cmidrule{2-7}
    & FedRec & 45822 & 877 & 3360 & 365 & 6720 (0.5x) \\
    & AsyncFedAvg & 30894 & 609 & 2357 & 255 & 4715 (0.7x) \\
    & FedAsync & 247068 & 4727 & 18165 & 1981 & 36331 (0.1x) \\
    \bottomrule
    \end{tabular}
\caption{\textbf{CIFAR-10 Performance Metrics on Homogeneous Cluster.} SemiSync with Momentum outperforms all other Synchronous and Semi-Synchronous policies. The first column refers to the federated learning domain and the target accuracy that each policy needs to reach. Column \textquotesingle Com. Cost \textquotesingle is an abbreviation for Communication Cost. The total models exchanged during training is twice the communication cost value. For every experiment the energy savings are computed against the synchronous Vanilla SGD baseline.}
\label{tbl:Cifar10HomogeneousCluster}
\end{table}

\begin{table}[htpb]
\noindent
\footnotesize
\setlength\tabcolsep{1.5pt}
    \centering
    \begin{tabular}{@{}clccccccccccc@{}}
    & &
    \multicolumn{3}{c}{\textbf{\begin{tabular}[c]{@{}c@{}}Fast (GPU)\end{tabular}}} & \multicolumn{3}{c}{\textbf{\begin{tabular}[c]{@{}c@{}}Slow (CPU)\end{tabular}}} &
    \multicolumn{5}{c}{\textbf{\begin{tabular}[c]{@{}c@{}}Total\end{tabular}}}
    \\
    \cmidrule(lr){3-5} \cmidrule(lr){6-8} \cmidrule(lr){9-13}
    \multicolumn{1}{c}{\textbf{\begin{tabular}[c]{@{}c@{}}Exp.\end{tabular}}} &
    \multicolumn{1}{c}{\textbf{\begin{tabular}[c]{@{}c@{}}Policy\end{tabular}}} &
    \multicolumn{1}{c}{\textbf{\begin{tabular}[c]{@{}c@{}}Iter.\end{tabular}}} &
    \multicolumn{1}{c}{\textbf{\begin{tabular}[c]{@{}c@{}}CT(s)\end{tabular}}} &
    \multicolumn{1}{c}{\textbf{\begin{tabular}[c]{@{}c@{}}EC\end{tabular}}} &
    \multicolumn{1}{c}{\textbf{\begin{tabular}[c]{@{}c@{}}Iter.\end{tabular}}} &
    \multicolumn{1}{c}{\textbf{\begin{tabular}[c]{@{}c@{}}CT(s)\end{tabular}}} &
    \multicolumn{1}{c}{\textbf{\begin{tabular}[c]{@{}c@{}}EC\end{tabular}}} &
    \multicolumn{1}{c}{\textbf{\begin{tabular}[c]{@{}c@{}}Iter.\end{tabular}}} &
    \multicolumn{1}{c}{\textbf{\begin{tabular}[c]{@{}c@{}}PT(s)\end{tabular}}} &
    \multicolumn{1}{c}{\textbf{\begin{tabular}[c]{@{}c@{}}CT(s)\end{tabular}}} &
    \multicolumn{1}{c}{\textbf{\begin{tabular}[c]{@{}c@{}}CC\end{tabular}}} &
    \multicolumn{1}{c}{\textbf{\begin{tabular}[c]{@{}c@{}}EC(EF)\end{tabular}}}
    \\
    \toprule
    \parbox[t]{7mm}{\multirow{10}{*}{\rotatebox[origin=c]{90}{\begin{tabular}[c]{@{}c@{}}Uniform \& IID\\0.75\end{tabular}}}}
    & Sync w/ Vanilla & 24000 & 578 & 1157 & 24000 & 16129 & 16129 & 48000 & 3225 & 16707 & 240 & 17286 \\
    & SemiSync ($\lambda=2$) w/ Vanilla & 45250 & 1018 & 2036 & 4750 & 3045 & 3045 & 50000 & 631 & 4064 & 100 & 5082 (3.4x)\\
    \cmidrule{2-13}
    & Sync w/ Momentum & 11000 & 185 & 371 & 11000 & 2700 & 2700 & 22000 & 540 & 2885 & 110 & 3071 (5.6x)\\
    & SemiSync ($\lambda=2$) w/ Momentum & 20250 & 335 & 671 & 2250 & 1217 & 1217 & 22500 & \textbf{269} & \textbf{1553} & \textbf{50} & \textbf{1889 (9.1x)} \\
    \cmidrule{2-13}
    & Sync w/ FedProx & 24000 & 548 & 1097 & 24000 & 22015 & 22015 & 48000 & 4403 & 22564 & 240 & 23113 (0.7x)\\
    & SemiSync ($\lambda=2$) w/ FedProx & 40250 & 817 & 1634 & 4250 & 4074 & 4074 & 44500 & 928 & 4891 & 90 & 5708 (3x)\\
    \cmidrule{2-13}
    & FedRec & 50800 & 1588 & 3177 & 1600 & 1449 & 1449 & 52400 & 732 & 3038 & 261 & 4627 (3.7x)\\
    & AsyncFedAvg & 69200 & 2439 & 4879 & 8800 & 2870 & 2870 & 78000 & 1315 & 5310 & 389 & 7750 (2.2x)\\
    & FedAsync & 76600 & 2703 & 5406 & 8000 & 3147 & 3147 & 84600 & 1406 & 5850 & 422 & 8554 (2x)\\
    \midrule
    \parbox[t]{7mm}{\multirow{10}{*}{\rotatebox[origin=c]{90}{\begin{tabular}[c]{@{}c@{}}Skewed \& Non-IID(5)\\0.65\end{tabular}}}}
    & Sync w/ Vanilla & 28200 & 1010 & 2020 & 22300 & 22812 & 22812 & 50500 & 4562 & 23822 & 250 & 24832\\
    & SemiSync ($\lambda=2$) w/ Vanilla	& 164082 & 3585 & 7170 & 16603 & 10006 & 10006 & 180685 & 2153 & 13591 & 220 & 17176 (1.4x)\\
    \cmidrule{2-13}
    & Sync w/ Momentum & 28200 & 670 & 1341 & 22300 & 10896 & 10896 & 50500 & 2179 & 11567 & 250 & 12238 (2x)\\
    & SemiSync ($\lambda=2$) w/ Momentum & 93882 & 1446 & 2893 & 9583 & 4434 & 4434 & 103465 & \textbf{1059} & \textbf{5881}  & \textbf{130} & \textbf{7328 (3.3x)}\\
    \cmidrule{2-13}
    & Sync w/ FedProx & 29328 & 1065 & 2130 & 23192 & 22998 & 22998 & 52520 & 4599 & 24063 & 260 & 25129 (0.9x)\\
    & SemiSync ($\lambda=2$) w/ FedProx & 210882 & 4677 & 9354 & 21283 & 12872 & 12872 & 232165 & 2884 & 17549 & 280 & 22226 (1.1x)\\
    \cmidrule{2-13}
    & FedRec & 155731 & 5205 & 10411 & 4378 & 4448 & 4448 & 160109 & 2431 & 9654 & 803 & 14860 (1.6x)\\
    & AsyncFedAvg & 183982 & 6185 & 12370 & 17361 & 7349 & 7349 & 201343 & 3346 & 13534 & 1021 & 19719 (1.2x)\\
    & FedAsync & \multicolumn{11}{c}{\textit{did not reach target accuracy}} \\
    \midrule
    \parbox[t]{7mm}{\multirow{11}{*}{\rotatebox[origin=c]{90}{\begin{tabular}[c]{@{}c@{}}Power Law \& Non-IID(3)\\0.6\end{tabular}}}}
    & Sync w/ Vanilla & 9664 & 600 & 1201 & 6496 & 10107 & 10107 & 16160 & 2021 & 10707 & 80 & 11308\\
    & SemiSync ($\lambda=2$) w/ Vanilla	& 80102 & 1831 & 3662 & 8183 & 4939 & 4939 & 88285 & 1103 & 6770 & 80 & 8601 (1.3x)\\
    \cmidrule{2-13}
    & Sync w/ Momentum & \multicolumn{11}{c}{\textit{did not reach target accuracy}} \\
    & SemiSync ($\lambda=2$) w/ Momentum & 34502 & 573 & 1147 & 3623 & 1723 & 1723 & 38125 & \textbf{438} & \textbf{2296}  & \textbf{40} & \textbf{2870 (3.9x)}\\
    \cmidrule{2-13}
    & Sync w/ FedProx & 13288 & 863 & 1727 & 8932 & 16363 & 16363 & 22220 & 3272 & 17227 & 110 & 18091 (0.6x)\\
    & SemiSync ($\lambda=2$) w/ FedProx & 102902 & 2084 & 4168 & 10463 & 7084 & 7084 & 113365 & 1719 & 9168 & 100 & 11252 (1x)\\
    \cmidrule{2-13}
    & FedRec & 90458 & 3671 & 7343 & 2647 & 2922 & 2922 & 93105 & 1832 & 6594 & 664 & 10266 (1.1x)\\
    & AsyncFedAvg & \multicolumn{11}{c}{\textit{did not reach target accuracy}} \\
    & FedAsync & \multicolumn{11}{c}{\textit{did not reach target accuracy}} \\     
    \bottomrule
    \end{tabular}
\caption{\textbf{CIFAR-10 Performance Metrics on Heterogeneous Cluster}. Semi-Synchronous with Momentum outperforms all other policies in time, communication, and energy costs. First column refers to the federated learning domain and the target accuracy that each policy needs to reach. (\textit{Iter.} = total number of local iterations, \textit{PT(s)} = parallel processing time in seconds, \textit{CT(s)} = cumulative processing time in seconds, \textit{CC} = communication cost, and \textit{EC(EF)} = energy cost with energy efficiency factor). For every experiment the energy savings are computed against the synchronous Vanilla SGD baseline.}
\label{tbl:Cifar10HeterogeneousCluster}
\end{table}

\paragraph{The Case for Federated Learning.}
Federated Learning is an efficient distributed machine learning paradigm that can allow multiple learning sites (learners) to jointly train a machine learning model without the need to share data (i.e., move the data out of their original source to a central site). This is critical in many domains, particularly in health informatics and in biomedical/genetics research consortia, where it is hard to share data due to privacy protections. 

To demonstrate the benefits of Federated Learning in a realistic biomedical domain, we analyzed the performance of individual sites training on their local data versus training in a federation for the BrainAge domain under different data distributions.
Figure~\ref{fig:Silos_vs_Federation}(a) considers an Uniform \& IID environment, where each site has the same amount of data and full representation of age ranges (cf. Figure\ref{fig:UKBB_DataDistributions}(a,b)). Even in this benign environment, the community model obtained by the federation significantly outperforms the model achieved by any single site. 
Figure~\ref{fig:Silos_vs_Federation}(b) considers a Skewed \& Non-IID environment, where different sites have different amounts of data and different age distributions (cf. Figure\ref{fig:UKBB_DataDistributions}(c,d)). This more realistic scenario would be typical of research consortia or federation of hospitals and clinics of different sizes and different disease prevalence. In this case, sites with smaller datasets or data distributions farther from the global distribution show very poor performance. 
Again, even the sites with larger datasets cannot match the community model. In summary, sites both small and large have strong incentives to join in Federated Learning.

\section{Discussion}
\label{sec:Discussion}
We have presented a novel Semi-Synchronous Federated Learning training policy, \textit{SemiSync}, with fast convergence and low communication and energy costs, particularly in heterogeneous data and computational environments. 
SemiSync defines a synchronization time point where all learners share their current local model to compute the community model. 
Compared to synchronous policies, learners with different amounts of data and/or computational power do not remain idle. By choosing the synchronization point so that the amount of data processed by the learners does not become too dissimilar, we can achieve faster convergence without increasing communication costs and with less energy consumption.
We performed extensive experiments comparing synchronous (FedAvg), asynchronous (FedAsync, FedRec), and semi-synchronous (SemiSync) policies in heterogeneous data and computational environments on standard benchmarks and on a challenging neuroimaging domain. We show experimentally that our SemiSync policy provides accelerated convergence and reaches better or comparable eventual accuracy than previous approaches.
The effects are more pronounced the more challenging the domain is, such as the results on CIFAR-100 and BrainAge with different data amounts per learner and Non-IID distributions. 

In future work, we plan to explore (1) adaptive hyperparameter (e.g., $\lambda$, $\eta$) schedules, (2) server-side model optimization update rules~\cite{reddi2020adaptive,wang2021field,hsu2019measuring}, and (3) federated learning policies under different homomorphic encryption schemes~\cite{stripelis2021:sipaim}. We also plan to adapt the SemiSync protocol to cross-device settings by considering connectivity and bandwidth limitations, and estimating hyperparameter lambda from a sample of learners and progressively fine-tuning it as more learners are sampled.

\begin{acks}
This research was supported in part by the Defense Advanced Research Projects Activity (DARPA) under contract HR0011\-2090104, and in part by the National Institutes of Health (NIH) under grants U01AG068057 and RF1AG051710.  The views and conclusions contained herein are those of the authors and should not be interpreted as necessarily representing the official policies or endorsements, either expressed or implied, of DARPA, NIH, or the U.S. Government. This is a study of previously collected, anonymized, de-identified data, available in a public repository. Data access approved by UK Biobank under Application Number 11559.
\end{acks}

\bibliographystyle{ACM-Reference-Format}
\bibliography{references}

\appendix
\section{Federated Problem Formulation}
\textbf{Global Loss.} Our learning environment consists of $N$ learners, each owning $n_k$ local training examples ($n_k = |D_k| = |(\mathcal{X}_k, \mathcal{Y}_k)|$), for a total number of $n$ examples across the federation, $n=\sum_k^N n_k$. Every learner computes its local objective, $F_k(w)$ by minimizing the empirical risk over its local training set $D_k^T$ as $F_k(w) = \mathbb{E}_{(x_k, y_k) \sim (\mathcal{X}_k, \mathcal{Y}_k)}[\ell_k(w;x_k, y_k)]$, with $\ell_k$ denoting the local loss function. The goal is to minimize the global function $f(w)$ to find the optimal set of parameters $w^*$ such that:
\begin{equation}\label{eq:FederatedGlobalLoss}
w^*=\underset{w}{\mathrm{argmin}} f(w) \quad\text{where}\quad f(w) = \sum_{k=1}^{N}\frac{n_k}{n}F_k(w) = \sum_{k=1}^{N}\frac{n_k}{n}\mathbb{E}_{(x_k, y_k) \sim (\mathcal{X}_k, \mathcal{Y}_k)}[\ell_k(w;x_k, y_k)]
\end{equation}
\textbf{Synchronization Period.} For every participating learner $k$, we denote as $t_k^{e}$ the computation time required to train the global model on its local dataset for a single epoch, $d_k$ the download transmission cost to receive the global model from the controller and $u_k$ the upload transmission cost to send its local model to the controller. Let $d_{max} = max(d_k, \forall k \in N)$, $d_{min} = min(d_k, \forall k \in N)$, $t_{max}^{e} = max(t_k^e, \forall k \in N)$, $t_{min} = min(t_k^e, \forall k \in N)$ and $u_{max} = max(u_k, \forall k \in N)$, $u_{min} = min(u_k, \forall k \in N)$. The computational (processing power) heterogeneity across all participating learners in the federation is defined as 
$h_t = \frac{t_{max}^e}{t_{min}^e}$, the download communication cost heterogeneity as 
$h_d = \frac{d_{max}}{d_{min}}$, and the upload communication cost heterogeneity as 
$h_u = \frac{u_{max}}{u_{min}}$. 
In our federated learning setting, where learners are servers located at geographically distributed datacenters with high speed inter-connectivity, we consider the communication cost to be negligible (and in addition $h_d \approx 1$,  $h_u \approx 1$). Therefore, we only analyze heterogeneity in terms of computational (training) cost (with $h_t > 1$). 
For instance, in our empirical evaluation, the heterogeneity between CPUs and GPUs in the CIFAR-100 domain and ResNet-50 model has $h_t = 33$. The semi-synchronous protocol delegates more training steps ($\mathcal{B}_k$) to computationally fast learners (i.e., $t_k^e \ll t_{max}^e$), which quickly improve the models, even when the slow learners / stragglers (i.e., $t_k^e = t_{max}^e$) cannot contribute much to the learning in the same period. By tuning the $\lambda$ hyperparameter, SemiSync balances the work done by fast and slow learners to accelerate the model convergence rate.

\section{Convergence Analysis}
In this section, we provide an analysis of the federated model performance by quantifying the weight divergence of the federated model to its centralized counterpart~\cite{zhao2018federated,yang2019federated}. The smaller the weight divergence the better the model performance (e.g., improved accuracy)
\\
\\
\textbf{Weight Divergence.} 
Following the theoretical analysis of~\cite{zhao2018federated,yang2019federated} we want to bound the federated model weight divergence computed using the semi-synchronous protocol compared to the centralized model. 
Consider a classification problem with $C$ classes that follows a joint probability distribution $p(x,y)$ where $x \in \mathcal{X}$ and $y \in C$. The probability of class $c$ in the dataset is
$p(y = c) = \sum_x{p(x,c)}$. 
Let function $\phi$ map the input $x$ to a probability vector $\overline{\nu}$ over the classes, with $\phi_c$ denoting the probability of $x$ belonging to the class $c$. The neural network implements the map $\phi$, which is parameterized over the weights $w$ of the network.

Let $w_{c,r}$ denote the community/global model of the federation at federation round $r$ and $w_z$ the centralized model. We define the population loss $\ell(w)$ though the cross-entropy loss:

\begin{equation*}
    \ell(w) = \mathbb{E}_{(x_k, y_k) \sim (\mathcal{X}_k, \mathcal{Y}_k)}[ \sum_{c=1}^C \mathds{1}_{y=c} \text{log} \phi_c(w;x)] = \sum_{c=1}^C p(y=c) \mathbb{E}_{x|y=c}[\text{log} \phi_c(w;x)]
\end{equation*}

In a centralized setting, we want to optimize the loss function by performing consecutive iterations/steps over the training dataset. The update rule (using SGD) for a single step is:

\begin{equation*}
    w_z = w_{z-1} - \eta \sum_{c=1}^C p(y=c) \nabla_W \mathbb{E}_{x|y=c}[\text{log} \phi_{c}(w;x)]
\end{equation*}

In a federated setting, where N learners participate at every round (full client participation), with every learner $k$ having its own unique local training dataset and data distribution, $D_i \cap D_j = \emptyset$, $p_i(y=c) \neq p_j(y=c), i \neq j$, $|D_k| = n_k, \text{and } n = \sum_{k=1}^{N}\frac{n_k}{n}$, the update rule for a global iteration (using SGD) after every learner performed a single (local) step is given by:

\begin{equation*}
    w_{c,r} = \sum_{k=1}^N \frac{n_k}{n} (w_{c,r-1} - \eta \sum_{c=1}^C p_k(y=c) \nabla_W \mathbb{E}_{x|y=c}[\text{log} \phi_{c}(w;x)])
\end{equation*}

\begin{proposition}
\label{prop:weight_divergence_proposition}
Given a classification problem with $C$ classes, a federation of $N$ learners,  with each learner $k$ owning $n_k$ training samples following a non-IID data distribution $p_k$, where $p_k \neq p_j, \forall k \neq j$. If  $\; \nabla_{w} \mathbb{E}_{x|y=c} \log f_c(x, w)$ is $\mathcal{M}_{x|y=c}$-Lipschitz for each class $c \in [C]$ and every learner synchronizes its local model every $\mathcal{B}_k$ local steps, then the weight divergence of SemiSync is bounded by:
\begin{align*}
||w_{c,r} - w_z|| &\leq \sum_{k=1}^{N} \frac{\mathcal{B}_{max}}{\mathcal{B}} \alpha_k^{\mathcal{B}_{max}}||w_{k,r-1}^{\mathcal{B}_{max}} - w_{z-1}^{\mathcal{B}_{max}}|| \\
&\qquad + \eta \sum_{c=1}^C ||p_k(y=c) - p(y=c)||(\sum_{j=0}^{\mathcal{B}_{max}-1}(\alpha_k)^j g_{max}(w_{z-1}^{\mathcal{B}_{max}-1-j} )))
\end{align*}
\end{proposition}

\begin{proof}
\label{prop:weight_divergence_proof} 
Let $w_{k,r}^{\mathcal{B}_k}$ denote the local model of learner $k$ at federation round $r$ after $\mathcal{B}_k$ local steps, $\mathcal{B} = \sum_k^N \mathcal{B}_k$ the total number of iterations across all learners and $w_z^{\mathcal{B}}$ the centralized model after $\mathcal{B}$ iterations. To measure weights divergence when learners perform disproportional number of steps during local training, we analyze the federation model convergence by weighting the local model of each learner on the total number of local steps it performed during training, 
i.e., $w_{c,r} = \sum_k^N\frac{\mathcal{B}_k}{\mathcal{B}} w_{k,r}^{\mathcal{B}_k}$:

\begin{align*}
||w_{c,r} - w_z|| &=
\sum_{k=1}^N ||\frac{\mathcal{B}_k}{\mathcal{B}} w_{k,r}^{\mathcal{B}_k} - w_z^{\mathcal{B}}|| 
\\
&\leq \sum_{k=1}^N \frac{\mathcal{B}_k}{\mathcal{B}} ||w_{k,r}^{\mathcal{B}_k} - w_z^{\mathcal{B}}|| 
\\
&= \sum_{k=1}^N\frac{\mathcal{B}_k}{\mathcal{B}} || w_{k,r}^{\mathcal{B}_k-1} -\eta \sum_{c=1}^C p_k(y=c)\nabla_{w}\mathbb{E}_{x|y=c} \log \phi_{c}(x, w_{k,r}^{\mathcal{B}_k-1}) 
\\ 
&\qquad - w_{z}^{\mathcal{B}-1} + \eta \sum_{c=1}^C p(y=c)\nabla_{w}\mathbb{E}_{x|y=c}\log \phi_{c}(x, w_{z}^{\mathcal{B}-1}) || 
\\
\end{align*}

\begin{align*}
\qquad\qquad\qquad\qquad\quad&\stackrel{(1)}{\leq} \sum_{k=1}^N\frac{\mathcal{B}_k}{\mathcal{B}} ||w_{k,r}^{\mathcal{B}_k-1} - w_{z}^{\mathcal{B}-1}|| 
\\
 &\qquad + \sum_{k=1}^N\frac{\mathcal{B}_k}{\mathcal{B}} \eta || \sum_{k=1}^N \frac{n_k}{n} \sum_{c=1}^C p_k(y=c) (\nabla_{w}\mathbb{E}_{x|y=c} \log \phi_{c}(x, w_{k,r}^{\mathcal{B}_k-1}) - \\
 &\qquad \qquad \qquad \qquad \qquad \qquad \qquad \qquad \nabla_{w}\mathbb{E}_{x|y=c}\log \phi_{c}(x, w_{z}^{\mathcal{B}-1}) || 
\\
 &\stackrel{(2)}{\leq} \sum_{k=1}^N\frac{\mathcal{B}_k}{\mathcal{B}} ||w_{k,r}^{\mathcal{B}_k-1} - w_{z}^{\mathcal{B}-1}|| \\
 &\qquad + \sum_{k=1}^N\frac{\mathcal{B}_k}{\mathcal{B}} \eta \sum_{k=1}^N \frac{n_k}{n} \sum_{c=1}^C p_k(y=c) \mathcal{M} ||w_{k,r}^{\mathcal{B}_k-1} - w_{z}^{\mathcal{B}-1}|| \\
 &= \sum_{k=1}^N\frac{\mathcal{B}_k}{\mathcal{B}} (1 + \eta\sum_{k=1}^N \frac{n_k}{n}\sum_{c=1}^C p_k(y=c) \mathcal{M}) ||w_{k,r}^{\mathcal{B}_k-1} - w_{z}^{\mathcal{B}-1}|| 
\\
&\stackrel{(3)}{=} \sum_{k=1}^{N} \frac{\mathcal{B}_k}{\mathcal{B}} \alpha_k ||w_{k,r}^{\mathcal{B}_k-1} - w_{z}^{\mathcal{B}-1}|| \numberthis \label{eq:weight_divergence_norm_1}
\end{align*}

Inequality (1) holds because the global dataset population for any class $c \in [C]$ is equal to the weighted average of the individual class populations within each learner, i.e., $p(y=c) = \sum_k^N \frac{n_k}{n}p_k(y=c)$. Inequality (2) holds because we assume $\nabla_{w} \mathbb{E}_{x|y=c} \log \phi_{c}(x, w)$ to be $\mathcal{M}_{x|y=c}$-Lipschitz. For equality (3) we set, $\alpha_k = (1 + \eta\sum_k^N \frac{n_k}{n}\sum_c^C p_k(y=c) \mathcal{M})$. From equation~\ref{eq:weight_divergence_norm_1}, if we assign to every learner the maximum number of local steps within a single round (i.e., the steps that the fastest learner in the federation performs), we get the following inequality for the divergence of the federated model from the centralized model, with $\sum_k^{N} \frac{\mathcal{B}_{max}}{\mathcal{B}} > 1$ and $\mathcal{B}_k \leq \mathcal{B}_{max}, \forall k \in [N]$:

\begin{align*}
||w_{c,r} - w_z|| \leq \sum_{k=1}^{N} \frac{\mathcal{B}_{max}}{\mathcal{B}} \alpha_k ||w_{k,r}^{\mathcal{B}_{max}-1} - w_{z}^{\mathcal{B}_{max}-1}|| \numberthis \label{eq:weight_divergence_norm_2}
\end{align*}

Based on the proposition 3.1 from the work of~\cite{zhao2018federated}, it holds that the weight divergence of the local model of any client $k \in [N]$ from the centralized model after $\mathcal{B}_{max}-1$ iterations is:

\begin{align*}
||w_{k,r}^{\mathcal{B}_{max}-1} - w_{z}^{\mathcal{B}_{max}-1}|| 
&\leq \alpha_k^{\mathcal{B}_{max}-1}||w_{k,r-1}^{\mathcal{B}_{max}} - w_{z-1}^{\mathcal{B}_{max}}|| \\
&\qquad + \eta \sum_{c=1}^C ||p_k(y=c) - p(y=c)||(\sum_{j=0}^{\mathcal{B}_{max}-2}(\alpha_k)^j g_{max}(w_{z}^{\mathcal{B}_{max}-2-j} )) \numberthis \label{eq:local_centralized_weight_divergence_norm}
\end{align*}
where $g_{max}$ is equal to:
\begin{align*}
 g_{max}(w_{z}^{\mathcal{B}_{max}-2}) = max_{c=1}^C || \nabla_{w}\mathbb{E}_{x|y=c}\log \phi_{c}(x, w_{z}^{\mathcal{B}_{max}-2}) ||
\end{align*}

By combining equations~\ref{eq:weight_divergence_norm_2} and~\ref{eq:local_centralized_weight_divergence_norm} we derive the following inequality:
\begin{align*}
||w_{c,r} - w_z|| &\leq \sum_{k=1}^{N} \frac{\mathcal{B}_{max}}{\mathcal{B}} (\alpha_k^{\mathcal{B}_{max}}||w_{k,r-1}^{\mathcal{B}_{max}} - w_{z-1}^{\mathcal{B}_{max}}|| \\
&\qquad + \eta \sum_{c=1}^C ||p_k(y=c) - p(y=c)||(\sum_{j=0}^{\mathcal{B}_{max}-1}(\alpha_k)^j g_{max}(w_{z-1}^{\mathcal{B}_{max}-1-j} )))
\end{align*}
Hence proved.
\end{proof}

\end{document}